%% file: root.tex
\documentclass[fleqn,10pt]{wlscirep}
\usepackage[utf8]{inputenc}
\usepackage[T1]{fontenc}
\usepackage[left]{lineno}
\usepackage{todonotes}
\usepackage{amsmath, amssymb}
\usepackage{mathtools}
\usepackage{comment}

\usepackage{array}
\newcommand{\PreserveBackslash}[1]{\let\temp=\\#1\let\\=\temp}
\newcolumntype{C}[1]{>{\PreserveBackslash\centering}p{#1}}
\newcolumntype{R}[1]{>{\PreserveBackslash\raggedleft}p{#1}}
\newcolumntype{L}[1]{>{\PreserveBackslash\raggedright}p{#1}}

\usepackage[most]{tcolorbox}

\usepackage{siunitx}
\usepackage{floatrow}
\usepackage{graphicx}
\usepackage[label font=bf,labelformat=simple]{subfig}
\usepackage[normalem]{ulem}
\usepackage{hyperref}

\usepackage{multirow}
\usepackage{fdsymbol}
\usepackage{bbding}
\usepackage{fontawesome5}

\usepackage{cleveref,tabularx}

\Crefname{figure}{Fig.}{Figs.}
\Crefname{section}{Sec.}{Secs.}
\Crefname{table}{Tab.}{Tabs.}
\Crefname{equation}{Equation}{Equations}

\newcommand{\rd}{\textit{redirection group}}
\newcommand{\ls}{\textit{leg swing group}}
\newcommand{\akf}{AKFI}
\newcommand{\pkf}{PKFI}

\newcommand{\coten}{\textrm{COT}_\textrm{en}}
\newcommand{\cotre}{\textrm{COT}_\textrm{re}}
\newcommand{\cotnr}{\textrm{COT}_\textrm{nr}}

\definecolor{rosa_d}{RGB}{138,0,82}
\definecolor{rosa_b}{RGB}{152, 114, 132}
\definecolor{cyan_d}{RGB}{0,138,138}
\definecolor{bronze}{RGB}{211, 130, 54}
\definecolor{gray}{RGB}{145, 149, 145}
\definecolor{sky_b}{RGB}{114, 166, 209}
\definecolor{sky}{RGB}{55, 126, 184}
\definecolor{sky_d}{RGB}{7, 90, 158}
\definecolor{green}{RGB}{0, 138, 55}

\graphicspath{{./figure/}}

\title{Passive knee flexion increases forward impulse of the trailing leg during the step-to-step transition}

\author[1,*]{Bernadett Kiss}
\author[2]{Alexandra Buchmann}
\author[2]{Daniel Renjewski}
\author[1,3]{Alexander Badri-Spröwitz}
\affil[1]{Max Planck Institute for Intelligent Systems, Stuttgart, 70569, Germany}
\affil[2]{Technical University of Munich, Chair of Applied Mechanics, TUM School of Engineering \& Design, Department of Mechanical Engineering, Garching near Munich, 85748, Germany}
\affil[3]{KU Leuven, Department of Mechanical Engineering, Leuven, 3000, Belgium}

\affil[*]{kiss@is.mpg.de}


\begin{abstract}
Human walking efficiency relies on the elastic recoil of the Achilles tendon, facilitated by a "catapult mechanism" that stores energy during stance and releases it during push-off. The catapult release mechanism could include the passive flexion of the knee, as the main part of knee flexion was reported to happen passively after leading leg touch-down. This study is the first to investigate the effects of passive versus active knee flexion initiation, using the bipedal EcoWalker-2 robot with passive ankles. By leveraging the precision of robotic measurements, we aimed to elucidate the importance of timing of gait events and its impact on momentum and kinetic energy changes of the robot. The EcoWalker-2 walked successfully with both initiation methods, maintaining toe clearance. Passive knee flexion initiation resulted in a 3\% of the gait cycle later onset of ankle plantar flexion, leading to 87\% larger increase in the trailing leg horizontal momentum, and 188\% larger magnitude increase in the center of mass momentum vector during the step-to-step transition. Our findings highlight the role of knee flexion in the release of the catapult, and timing of gait events, providing insights into human-like walking mechanics and potential applications in rehabilitation, orthosis, and prosthesis development.
\end{abstract}

\begin{document}

\begin{titlepage}

\centering
This preprint has been submitted for publication in the Biomimetics Collection of Scientific Reports. \\
\vspace{1cm}
Manuscript submitted: 29 October 2024 \\
\vspace{1cm}
Biomimetics Collection of Scientific Reports: \url{https://www.nature.com/collections/adjghffbah} \\

\pagestyle{empty}

\end{titlepage}

\flushbottom
\maketitle

\pagestyle{plain} 


\include{Sections/Introduction}
\include{Sections/Results}
\include{Sections/Discussion}
\include{Sections/Methods}

\bibliography{root}
\section*{Data availability}

All data needed to evaluate the conclusions of the paper are available in the paper or the Supplementary Materials. CAD design files of the EcoWalker-2 robot, control code, data analysis and visualization code, and experimental data are available for non-commercial use at \url{ https://doi.org/10.17617/3.BJ584M}. Video footage of the EcoWalker-2 robot's locomotion on the treadmill in both AKFI and PKFI experiments is available at \url{https://www.youtube.com/watch?v=RupuZPBI6Bg} and at \url{https://www.youtube.com/watch?v=oWwJbTPUOM4}.

\section*{Acknowledgements}
The authors thank the International Max Planck Research School for Intelligent Systems (IMPRS-IS) for supporting Bernadett Kiss. The authors would like to thank Felix Grimminger and his team for the help with the actuator modules, the Robotics ZWE at MPI-IS for the 3D-printed parts, and Guido Nafz from the MPI-IS Precision Mechanics Workshop for the precision parts.

\section*{Author contributions statement}
B.K., A BS., A.B., and D.R. all contributed to the conceptual design of the study, 
B.K. modified the robot design and control, performed the experiments, conducted the data analysis, and data visualization under the supervision of A.BS, 
B.K. and A.BS. interpreted the results, 
B.K. wrote the original draft of the manuscript, 
A.BS., A.B., and D.R. contributed to the review and editing of the manuscript. 

\section*{Funding}
This work was funded by the Deutsche Forschungsgemeinschaft (DFG, German Research Foundation) - 449427815.

\section*{Additional information}
\textbf{Competing interests} The author(s) declare no competing interests.
\end{document}

%% file: Sections/Introduction.tex
\section*{Introduction}
%
%
%
Human walking gait emerges from the intricate interplay between neural motor control, reflexes, muscle dynamics, and passive structures. Notably, ankle mechanics play a crucial role, with the ankle plantar flexor muscles contributing up to \SI{78}{\%} of the total positive muscle work during push-off \cite{Meinders1998}. Moreover, the elastic recoil of the mechanically passive Achilles tendon is responsible for up to \SI{91}{\%} of the ankle power output during push-off \cite{Cronin2013}. A catapult mechanism in the human leg was proposed to generate the high power output at the ankle by gradually storing energy during stance and then quickly releasing the energy during push-off \cite{Hof1983, Lipfert2014}. A possible catapult release mechanism could include the passive flexion of the knee joint, as the main part of knee flexion was reported to happen passively after leading leg touch-down \cite{Perry1992}. Building on Perry's observation, we investigate the effects of passive versus active knee flexion initiation on gait events around push-off using a robotic system. By leveraging the precision of robotic measurements, we aim to elucidate the importance of timing of gait events and its impact on momentum and kinetic energy changes in the trailing leg, remaining body, and center of mass. Understanding the biological mechanisms driving natural leg dynamics during push-off is essential for advancing gait rehabilitation \cite{Kapsalyamov2024}, prosthesis \cite{tran2022}, orthosis, and exoskeleton development \cite{molinaro2024}, as well as efficient legged robot design \cite{Ames2014}. \\
%
%
%
The principles of passive dynamics underlying human gait have been extensively investigated with robots \cite{Mcgeer1990, Collins2005}. However, none of the passively walking bipedal robots based on passive dynamics have been studying the human lower-leg catapult function specifically. Instead, they were designed with other goals in mind; the Cornell biped was designed for minimal cost of transport, the MIT learning biped was designed to test motor learning utility on a passive-dynamic mechanical design, and the Delft biped was designed to test skateboard-like ankle joints for lateral stability \cite{Collins2005}. \\
The EcoWalker-2 robot, a new version of the EcoWalker robot \cite{Kiss2022}, in contrast to the crouched gait of many humanoids is capable of straight-leg walking gait. Straight-leg walking is more human-like and it improves energy efficiency as the required torque and energy consumption to support the body weight at the knee joints is lower than during bent-knee walking \cite{Winter2009, Kurazume2005}. A robot walking with straight legs in midstance can achieve greater ground clearance, enabling it to navigate over larger obstacles and reduce the risk of collisions, while also reducing the necessary range of motion and actuator velocities required to maintain a given walking speed \cite{Griffin2018}. \\
\begin{figure*}[!ht]
\centering
\sidesubfloat[]{\includegraphics[width=0.46\textwidth]{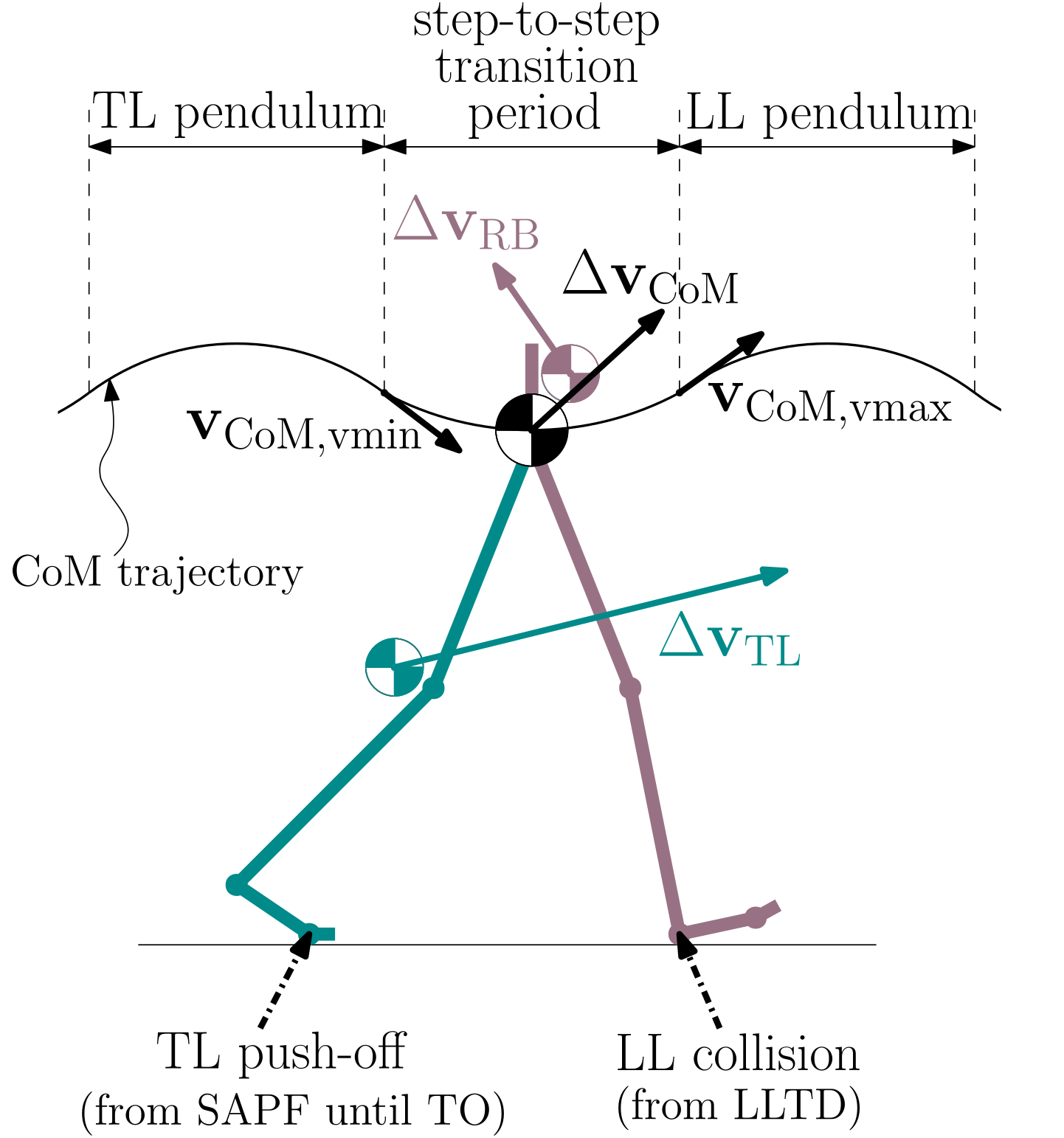}\label{fig:intro_views}}
\hfil
\sidesubfloat[]{\includegraphics[width=0.48\textwidth]{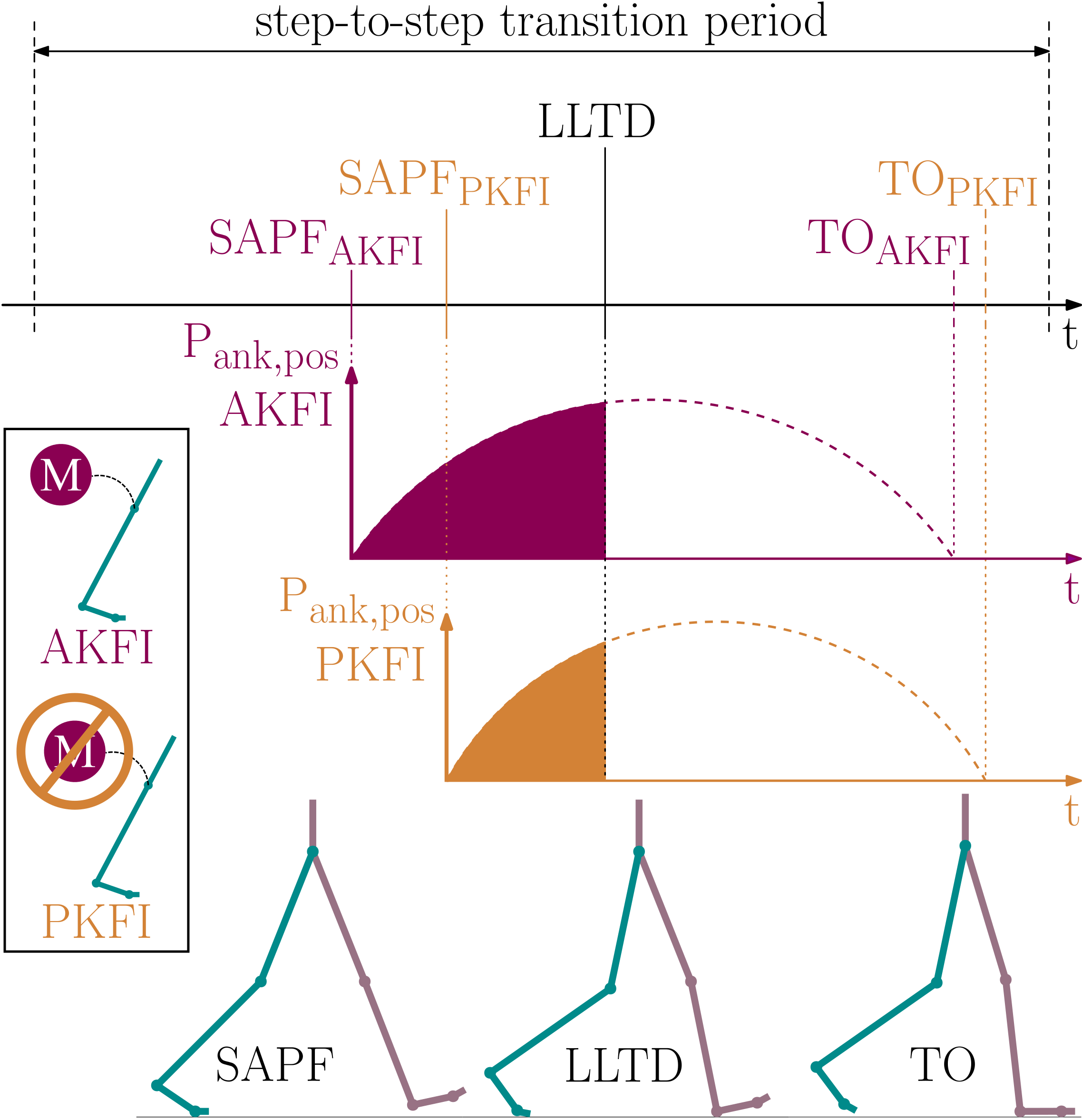}\label{fig:intro_timing_graph}}
\caption{The two main roles of push-off (\textbf{a}) and the expected effect of passive versus active knee flexion initiation (\textbf{b}). 
The main part of knee flexion, from 5-\SI{40}{deg}, occurs passively after LLTD \cite{Perry1992}. We compare \textcolor{bronze}{\pkf{}} and \textcolor{rosa_d}{\akf{}} to influence energy flow into the \textcolor{rosa_b}{RB} and \textcolor{cyan_d}{TL} by indirectly manipulating SAPF timing. We aim to contribute to the debate on the primary function of ankle push-off in the human swing leg catapult mechanism \cite{Zelik2016}.
\textbf{a}: During a step-to-step transition, push-off plays two main roles: it redirects the center of mass velocity ($\Delta \mathbf{v}_\mathrm{CoM}$) from one inverted pendulum arc to the next, and it increases the velocity of the \textcolor{cyan_d}{TL}. The drawing was inspired by Fig 1 in \cite{Adamczyk2009} and Fig 4 in \cite{Lipfert2014}.
\textbf{b}: We designed the \textcolor{rosa_d}{\akf{}} experiment with earlier knee flexion onset than \textcolor{bronze}{\pkf{}} experiment to alter the SAPF timing. This results in an earlier SAPF in \textcolor{rosa_d}{\akf{}}, prolonging the positive ankle power ($\mathrm{P}_\mathrm{ank, pos}$) period until LLTD. Before LLTD, the \textcolor{cyan_d}{TL} bears the whole body weight, while the loading leg (LL) is in swing. The filled part under the positive ankle power curve represents the predicted ankle energy mainly accelerating the \textcolor{rosa_b}{RB}. Due to the predicted SAPF delay in \textcolor{bronze}{\pkf{}} experiment, we expect lower ankle energy accelerating the \textcolor{rosa_b}{RB} in \textcolor{bronze}{\pkf{}} than in \textcolor{rosa_d}{\akf{}} experiment.\\
\textbf{Abbreviations:} \textcolor{cyan_d}{TL: Trailing Leg}, LL: Leading Leg, \textcolor{rosa_b}{RB: Remaining Body}, CoM: Center of Mass, vmin: time of minimum vertical velocity of the CoM (start of Step-to-Step transition period), vmax: second vertical velocity peak of the CoM after vmin (end of Step-to-Step transition period), SAPF: Start of Ankle Plantar Flexion, LLTD: Leading Leg Touch-Down, TO: Toe-Off, \textcolor{rosa_d}{\akf{}: Active Knee Flexion Initiation}, \textcolor{bronze}{\pkf{}: Passive Knee Flexion Initiation}.}
\label{fig:intro}
\end{figure*}
%
%
%
%
%
To achieve efficient walking, the EcoWalker-2 robot draws inspiration from the human lower leg's unique mechanical properties. In particular, the human lower leg can be thought of as a catapult system, where elastic energy is stored and rapidly released to facilitate movement. A catapult has three main mechanical components: an elastic element, a block, and a catch. In the human lower leg, it is challenging to identify the specific catapult mechanics and their functional elements due to the complex interplay between the thigh-shank-foot segment chain and muscle-tendon units at the ankle joint. Elastic energy storage and rapid recoil during ankle push-off are facilitated by the Achilles tendon, which is attached to the Soleus (SOL) and Gastrocnemius (GAS) muscles \cite{Fukunaga.2001,Ishikawa.2005,Lichtwark.2006,Lichtwark.2007}. The GAS muscle's mostly isometric work facilitates energy-efficient loading of the Achilles tendon, while the SOL muscle contributes to ankle coordination through active contraction \cite{Cronin.2013c}. The EcoWalker robot was used to demonstrate the benefits of a SOL spring-tendon, which amplified ankle power and enabled more efficient and faster walking \cite{Kiss2022}. In contrast, a GAS spring-tendon was shown to modify the coordination between the ankle and knee joints during ankle push-off \cite{Kiss2022}.
The block, serving as the supporting structure of the catapult, was determined to be a combination of the foot and the ground \cite{Renjewski2022}. The catch, a crucial component that locks and releases the human leg catapult, remains unidentified. \\
During the release of the catapult, the knee and the hip start to flex, and the ankle starts to plantarflex. A possible release mechanism can be derives from Perry's work \cite{Perry1992}. The main part of knee flexion, from \SI{5}{\degree} to \SI{40}{\degree}, was reported to happen passively after leading leg touch-down, as weight transfer to the leading leg frees the trailing leg's knee joint to flex passively \cite{Perry1992} (p. 104f). \\
In this work and for the first time, we test Perry's passive knee flexion process on a bipedal robot, as part of the lock-and-release mechanism of the human leg catapult. We test if passive knee flexion resulting from leg dynamics can trigger the human leg catapult release on a bipedal robot. We compare the same robotic system, the EcoWalker-2 robot \cite{Kiss2022}, with and without active knee flexion initiation: active knee flexion initiation (\akf{}) versus passive knee flexion initiation (\pkf{}). \\
In the \pkf{} experiment, the trailing leg's knee motor does not apply torque in the EcoWalker-2 robot \cite{Kiss2022}, the knee joint can move freely, from midstance until the next touch-down of the trailing leg. In the meantime, passive SOL and GAS spring-tendons provide plantar flexion and knee flexion moments. For sufficient toe clearance, knee flexion must happen before the start of the swing. As the knee motors will not provide knee flexion moment in the \pkf{} experiment, we expect that knee flexion will be a result of knee flexion moment from the GAS spring-tendon and knee flexion moment from the ground reaction force. \\
Body weight plays a crucial role in the loading of the ankle \cite{Renjewski2022}, and knee flexion induces an ankle alleviation \cite{Lipfert2014}, which in turn enables the onset of ankle plantarflexion. In the \akf{} experiment, knee flexion onset timing is expected to influence the timing of the start of ankle plantar flexion. The start of ankle plantar flexion is influenced by the active hip and active knee flexion that pull the foot off the ground during the \akf{} experiment. At the start of ankle plantar flexion the ankle power becomes positive, indicating the start of ankle unloading. Ankle unloading, the positive ankle power period, lasts until toe-off (\Cref{fig:intro_timing_graph}). \\
In the \akf{} experiment, knee flexion is actively initiated as the knee motors are commanded to follow a predefined trajectory during the whole gait cycle. We define the onset of knee flexion in the desired knee angle curve earlier than in the \pkf{} experiment. Due to the earlier timing of knee flexion onset, the start of ankle plantar flexion is also expected earlier in the \akf{} experiment than in the \pkf{} experiment. \\
To reach a symmetric gait with the bipedal EcoWalker-2 robot, the phase shift in the control of the hip motors is a constant \SI{50}{\%} of the gait cycle. As the phase shift is constant, the time of leading leg touch-down relative to trailing leg touch-down is not expected to largely change between \akf{} and \pkf{} experiments. Leading leg touch-down starts the weight transfer from the trailing leg to the leading leg, the trailing leg is freed up to go into swing during the double support phase. \\ 
Consequently, the time period between the start of ankle plantar flexion and leading leg touch-down is expected to be shorter in \pkf{} experiment than in \akf{} experiment (\Cref{fig:intro_timing_graph}).
By manipulating the timing of the start of ankle plantar flexion, we intend to influence the ratio of energy flow into the remaining body and the trailing leg. \\
%
%
Accurately studying the effects of such a gait event timing change can be challenging due to the inherent variability in human gait data, where the average standard deviation of gait events can be as high as \SI{1.5}{\%} of the gait cycle \cite{VanCriekinge2023}. In contrast, a robotic system can demonstrate greater repeatability, with an average standard deviation of less than \SI{0.2}{\%} of the gait cycle for gait events (see Supplementary Table S6). The almost tenfold better precision of robotic measurements allowed us to investigate the crucial role of timing of gait events and its influence on momentum and kinetic energy changes in the trailing leg, remaining body, and center of mass. \\
%
%
%
Our investigation into the effects of timing of gait events has wide-ranging implications for understanding the main function of push-off, a topic of ongoing debate in the field. Some researchers (\ls) argue that push-off energy primarily accelerates the swing leg, while others (\rd) propose that it aids in redirecting the center of mass velocity to reduce collision losses \cite{Zelik2016} (\Cref{fig:intro_views}). However, despite the extensive discussion on the function of push-off, the influence of gait event timing on its function has remained largely unexplored. By examining the factors that influence the push-off energy flow, such as the timing of ankle plantar flexion, we expect to gain a deeper understanding of the function of push-off in gait and contribute to resolving the long-standing debate. \\
The \ls{} of researchers see the main role of ankle push-off in its contribution to leg swing acceleration \cite{Winter1978, Meinders1998, Hof1992, Lipfert2014} (\Cref{fig:intro}). Segmental energy flow analysis on human walking data showed that the increase in the trailing leg energy is \SI{90}{\%} of the positive ankle energy generated by the plantarflexor muscles during push-off \cite{Meinders1998}. Meanwhile the energy increase of the trunk during push-off is \SI{13}{\%} of the positive ankle push-off energy \cite{Meinders1998}. \\
In human walking, each step-to-step transition involves the redirection of the center of mass (CoM) velocity from one inverted pendulum arc to the next \cite{Donelan2002_simPosNeg, Adamczyk2009} (\Cref{fig:intro_views}). The step-to-step transition period, defined as the interval from minimum to maximum vertical center of mass velocity, was used to capture the effect of the transition from trailing leg single support to leading leg single support \cite{Adamczyk2009} (\Cref{fig:intro}). The \rd{} of researchers suggest that, during the step-to-step transition, the main function of ankle push-off is aiding the redirection of the CoM velocity \cite{Kuo2002, Donelan2002_simPosNeg, Donelan2002_mechWork, Adamczyk2009, Soo2010, Lee2011, Lee2013, Zelik.2014, Faraji2018, Croft2020}. Experimental human gait data at \SI{0.9}{m/s} to \SI{1.8}{m/s} walking velocities shows that CoM energy change during push-off is \SI{86}{\%}-\SI{96}{\%} of ankle push-off energy \cite{Zelik.2010}. The \rd{} simulated human gait with simple models with or without a knee joint, and considered the body as a whole. The physics-based models predict that collision losses can be reduced by pushing off before leading leg touch-down \cite{Kuo2002, Donelan2002_mechWork,Donelan2002_simPosNeg, Ruina2005, Adamczyk2009, Zelik.2014, Bregman2011}. However, Donelan et al. point out that their calculation method cannot differentiate between the effects of swing leg motion and CoM redirection \cite{Donelan2002_mechWork,Donelan2002_simPosNeg}. \\
The combined idea of both \rd{} and \ls{} were already considered in biomechanics research in 1966 \cite{Inman1966}. Later Zelik and Adamczyk \cite{Zelik2016} supported the view that ankle push-off propels both the trailing leg into swing and the body over the leading leg. The ankle push-off energy contributes to leg swing acceleration, i.e. kinetic energy increases in the swing leg, but only a small part of the kinetic energy is transferred to the torso via the hip \cite{Meinders1998, Lipfert2014, Zelik2016}. However, as the swing leg is part of the CoM calculations, the swing leg acceleration does increase the CoM velocity, so most of the kinetic energy change in the swing leg also appears as a CoM kinetic energy change \cite{Zelik2016} (\Cref{fig:intro_views}). Therefore, Zelik and Adamczyk \cite{Zelik2016} suggest that the consideration of ankle mechanics should avoid the binary contrast of leg swing versus CoM redirection. Instead, in the interpretation of the function of the ankle push-off Zelik and Adamczyk \cite{Zelik2016} propose taking into account both swing leg acceleration and CoM redirection effects simultaneously. \\
To further illustrate this concept, let us consider the effect of trailing leg velocity change on the whole body CoM velocity change. By modeling the body as a system of segments, divided into the trailing leg and the remaining body, we can calculate the contribution of each segment to the whole body CoM momentum (\Cref{fig:intro_views}). The remaining body consists of the leading leg and the head-arms-trunk segments. Notably, the trailing leg's contribution to the whole body CoM momentum can be substantial, despite its smaller mass, if its velocity is high \cite{Lipfert2014, Zelik2016}. This is evident in experimental human gait data at the preferred transition speed between walking and running, where the trailing leg's horizontal momentum increased by \SI{22.2}{Ns} during push-off, while the remaining body's horizontal momentum decreased by \SI{7.8}{Ns}. Although the trailing leg's mass (\SI{11.4}{kg}) is only about one-sixth that of the remaining body (\SI{59.5}{kg}), its large velocity change resulted in a net increase in the whole body CoM's horizontal momentum by \SI{14.4}{Ns} \cite{Lipfert2014}. \\
%
%
%
During push-off, the momentum changes of the trailing leg and the remaining body depend on how much of the ankle push-off energy is transferred through the hip. Energy transfer during push-off could be influenced by the body weight distribution between the trailing and the leading leg. Specifically, when the trailing leg bears the body's weight during the single support phase, we anticipate that a larger proportion of the ankle energy in the trailing leg will contribute to accelerating the rest of the body. As the leading leg touches down and becomes the weight-bearing leg, we expect the trailing leg's ankle energy to increasingly accelerate the trailing leg into its swing phase, rather than the rest of the body (\Cref{fig:intro_timing_graph}). Hence, we expect that the exact timing of the start of ankle plantar flexion and the leading leg touch-down largely alters the step-to-step transition dynamics. \\
In \akf{} experiment, the longer duration between the start of ankle plantar flexion and leading leg touch-down is expected to result in a greater proportion of ankle energy being transferred to accelerate the remaining body, and a lesser proportion being used to accelerate the trailing leg during the step-to-step transition period, compared to \pkf{} experiment (\Cref{fig:intro_timing_graph}). We hypothesize that the momentum of the remaining body ($\textbf{p}_\textrm{RB}$) increases more and the momentum of the trailing leg ($\textbf{p}_\textrm{TL}$) increases less during the step-to-step transition in \akf{} experiment compared to \pkf{} experiment.

%% file: Sections/Results.tex
\section{Results}
\begin{figure*}[!ht]
\centering
\sidesubfloat[]{\includegraphics[width=0.48\linewidth]{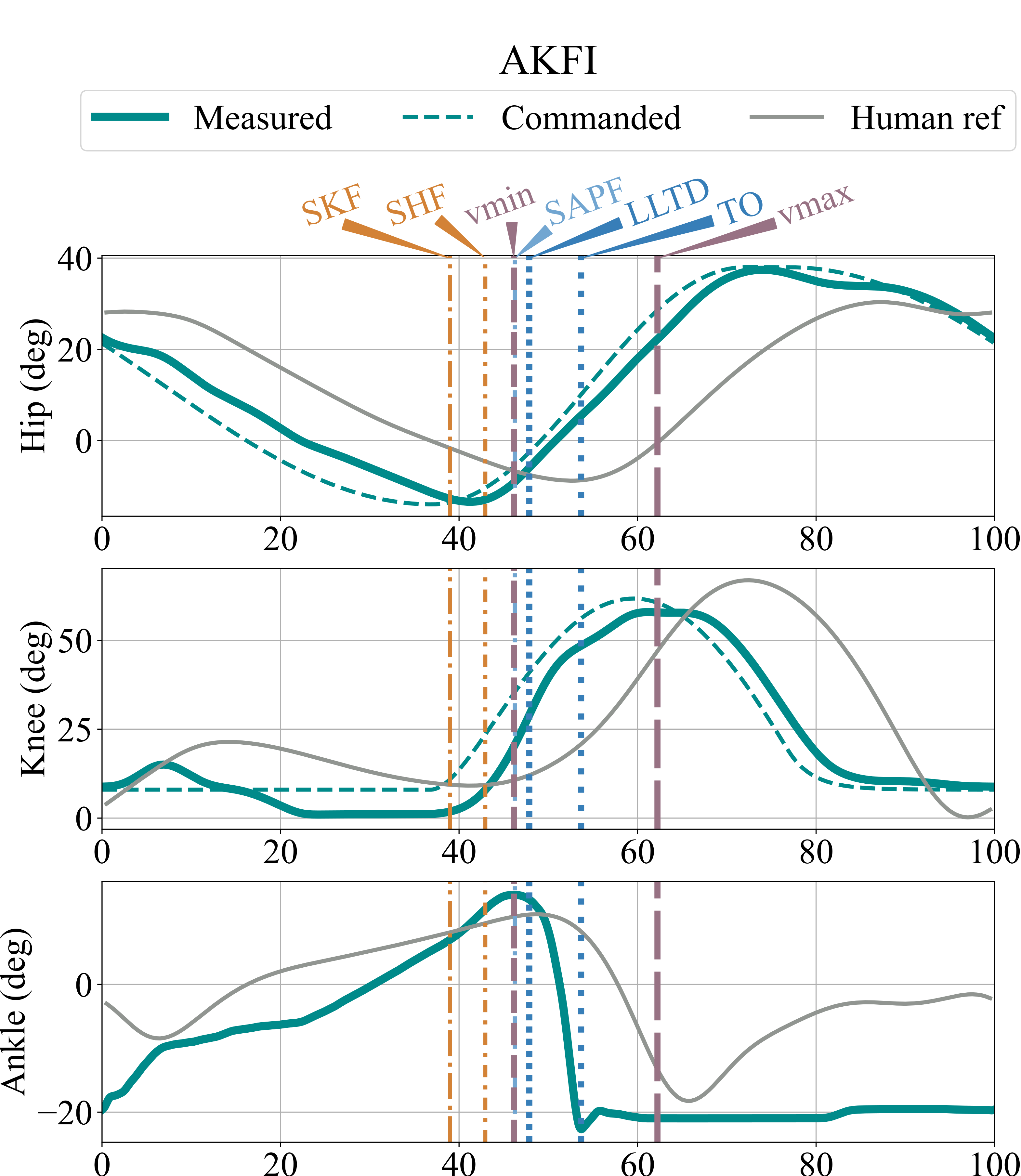}\label{fig:joint_angles:a}}
\hfil
\sidesubfloat[]{\includegraphics[width=0.48\linewidth]{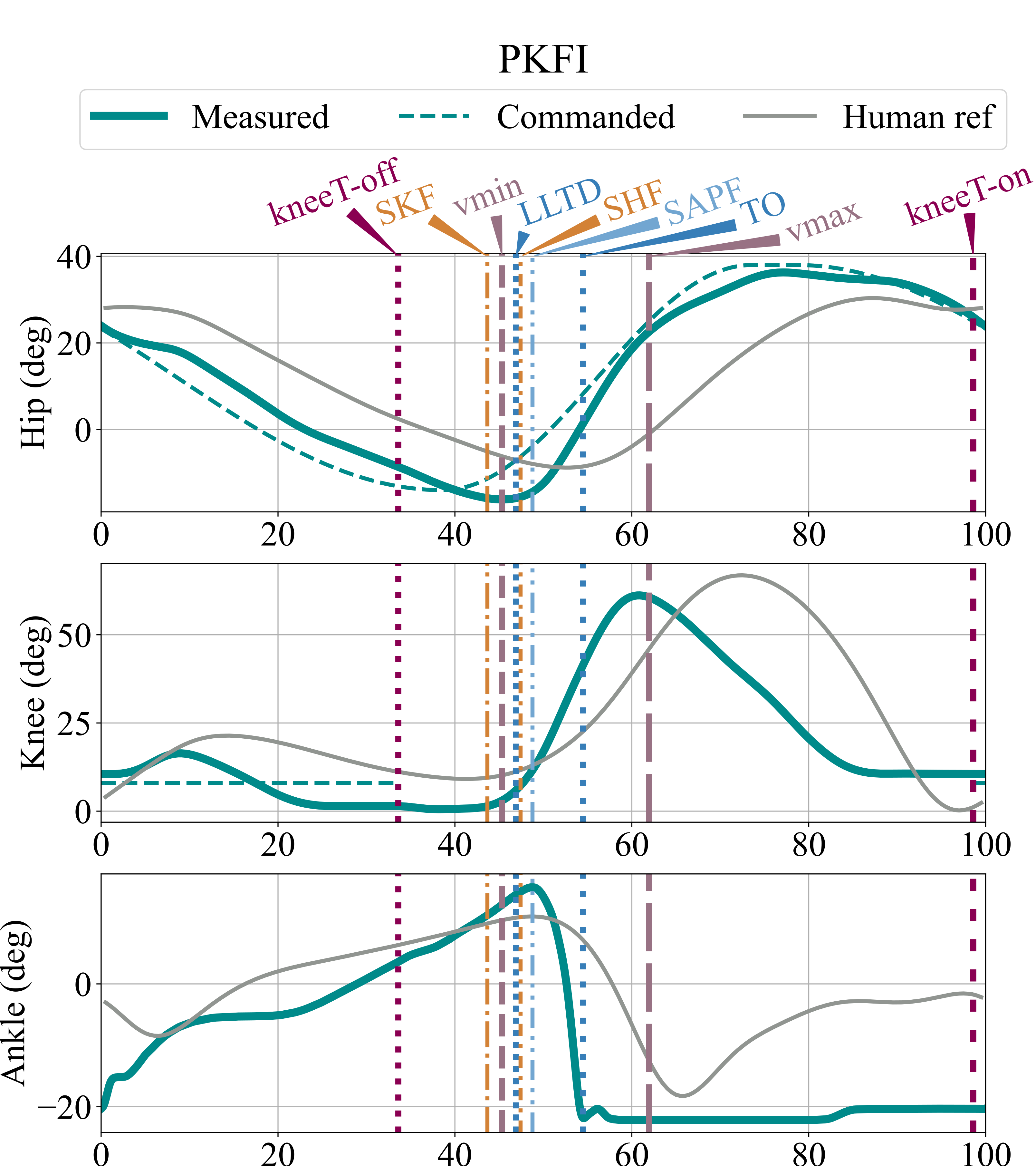}\label{fig:joint_angles:b}}
\caption{Hip, knee, and ankle angles during the full gait cycle in the experiments with active knee flexion initiation (\akf{} - \textbf{a}), and with passive knee flexion initiation (\pkf{} - \textbf{b}). Shading shows the standard deviation of the curves. Horizontal axis shows the gait cycle percentage, \SI{0}{\% GC} is the touch-down of the trailing leg. In the \pkf{} experiment, the knee motor torque is zero from the dark pink vertical dotted line (\textcolor{rosa_d}{kneeT-off}) until the dark pink vertical dashed line (\textcolor{rosa_d}{kneeT-on}). The continuous cyan line shows the \textcolor{cyan_d}{measured joint angles} of the robot, while the dashed cyan line shows the \textcolor{cyan_d}{commanded joint angles} of the hip and the knee. Joint angles of \textcolor{gray}{human walking} are overlayed (gray lines) for reference. \cite{VanderZee2022}$^\text{: average of trials 20, 21, and 22}$ \\ 
\textbf{Abbreviations:} \textcolor{bronze}{SKF:} Start of Knee Flexion, \textcolor{bronze}{SHF:} Start of Hip Flexion, \textcolor{sky_b}{SAPF:} Start of Ankle Plantar Flexion, \textcolor{rosa_b}{vmin:} time of minimum vertical velocity of the CoM, \textcolor{sky}{LLTD:} Leading Leg Touch-Down, \textcolor{sky}{TO:} Toe-Off, \textcolor{rosa_b}{vmax:} second vertical velocity peak of the CoM after vmin. }
\label{fig:joint_angles}
\end{figure*}

\begin{figure*}[ht]
\centering
\includegraphics[width=0.7\linewidth]{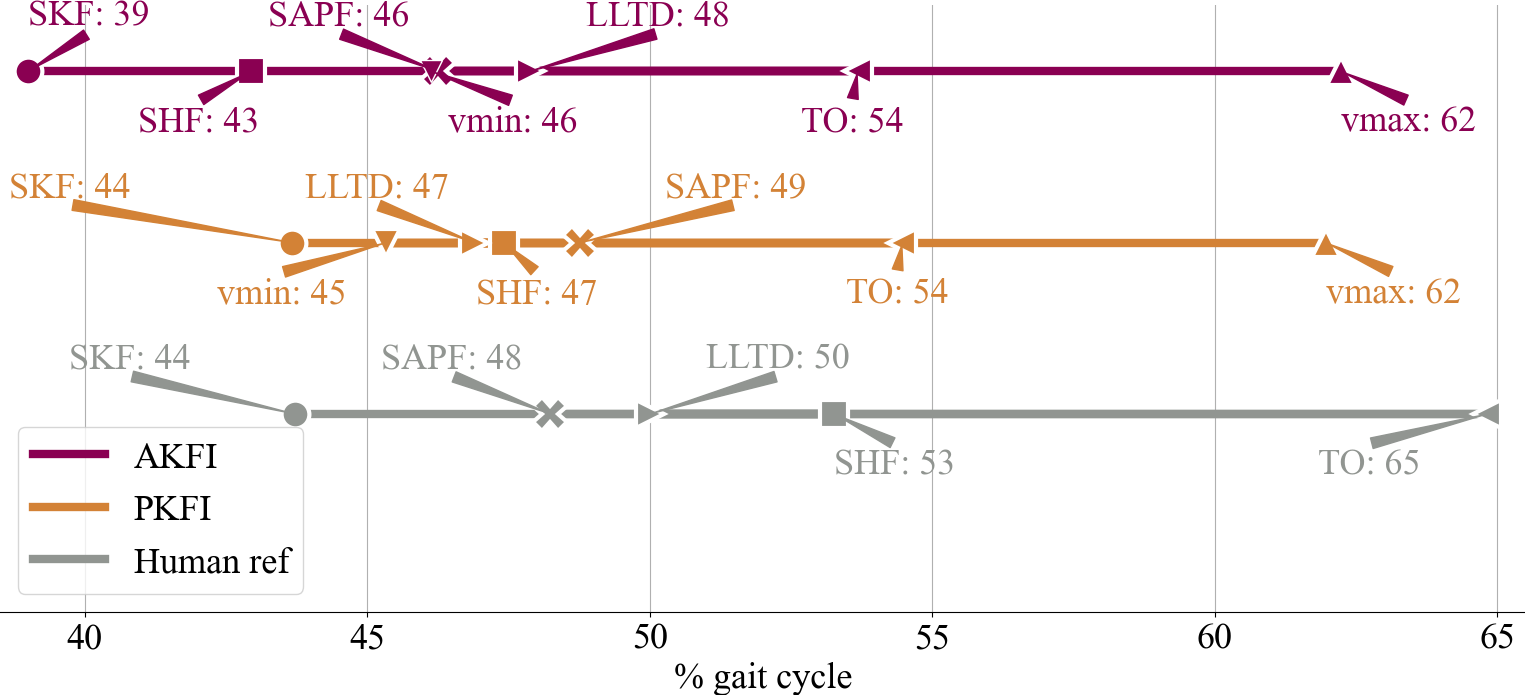}
\caption{Timing of the gait events in gait cycle percentage with \textcolor{rosa_d}{active knee flexion initiation (\akf{})}, and with \textcolor{bronze}{passive knee flexion initiation (\pkf{})}, and by \textcolor{gray}{humans} \cite{VanderZee2022}. \SI{0}{\% GC} is the touch-down of the trailing leg. In the \textcolor{bronze}{\pkf{}} experiment, knee and hip flexion start \SI{5}{\% GC} later than in the \textcolor{rosa_d}{\akf{}} experiment (SKF and SHF). LLTD occurs \SI{1}{\%GC} earlier with \textcolor{bronze}{\pkf{}} than with \textcolor{rosa_d}{\akf{}}. The ankle starts to plantarflex (SAPF) \SI{2}{\% GC} after LLTD with \textcolor{bronze}{\pkf{}}, while SAPF occurs \SI{2}{\% GC} before LLTD with \textcolor{rosa_d}{\akf{}}. The gait event timing values and their standard deviation values are available in Supplementary Table S2. \\ 
\textbf{Abbreviations:} $\mdlgblkcircle$ SKF: Start of Knee Flexion, $\mdlgblksquare$ SHF: Start of Hip Flexion, \XSolidBold SAPF: Start of Ankle Plantar Flexion, $\blacktriangledown$ vmin: time of minimum vertical velocity of the CoM, $\blacktriangleright$ LLTD: Leading Leg Touch-Down, $\blacktriangleleft$ TO: Toe-Off, $\blacktriangle$ vmax: second vertical velocity peak of the CoM after vmin.}
\label{fig:timing_graph}
\end{figure*}

\begin{figure*}[ht]
\centering
\sidesubfloat[]{\includegraphics[width=0.48\textwidth]{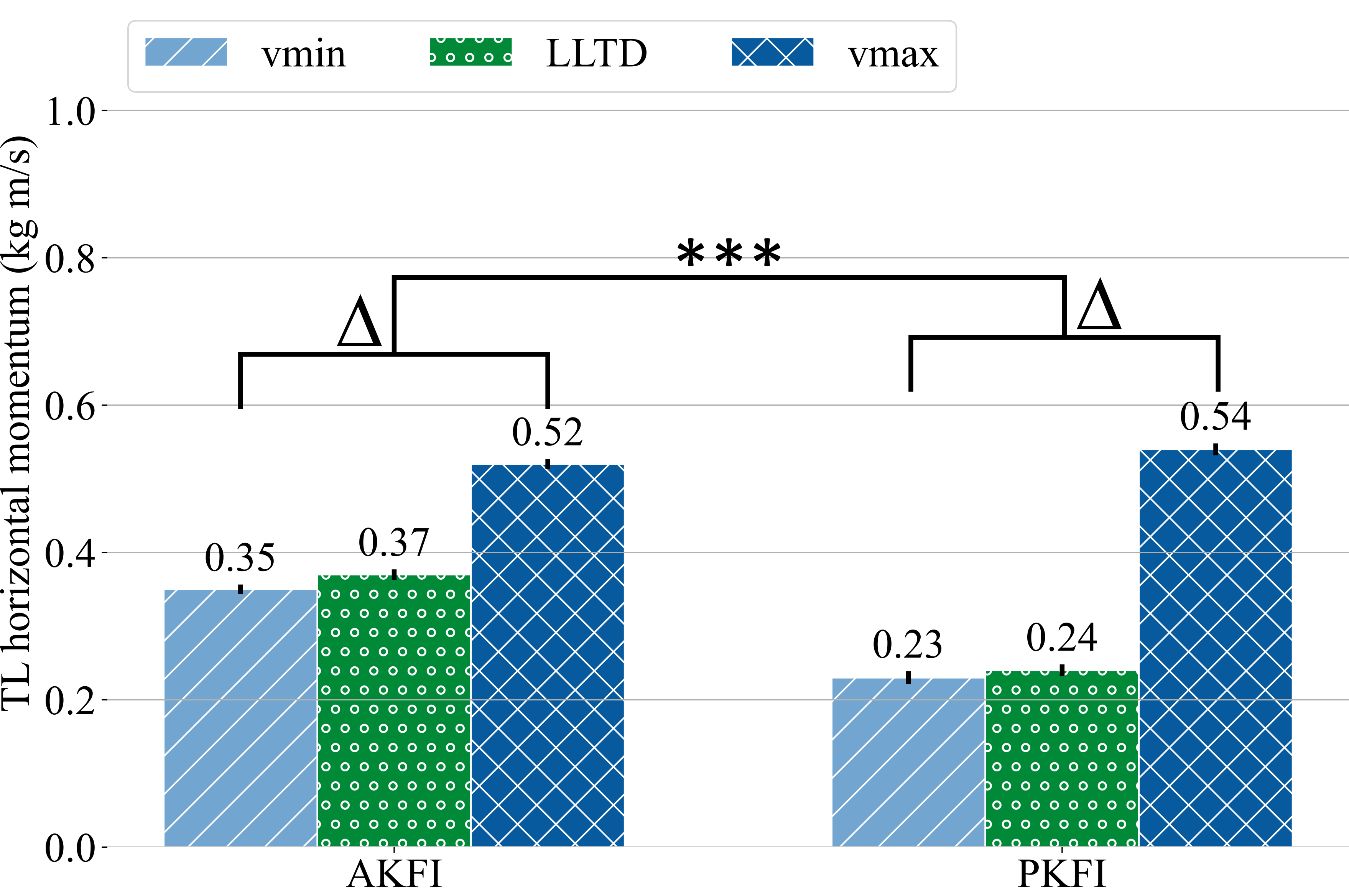}\label{fig:momentums:a}}
\hfil
\sidesubfloat[]{\includegraphics[width=0.48\textwidth]{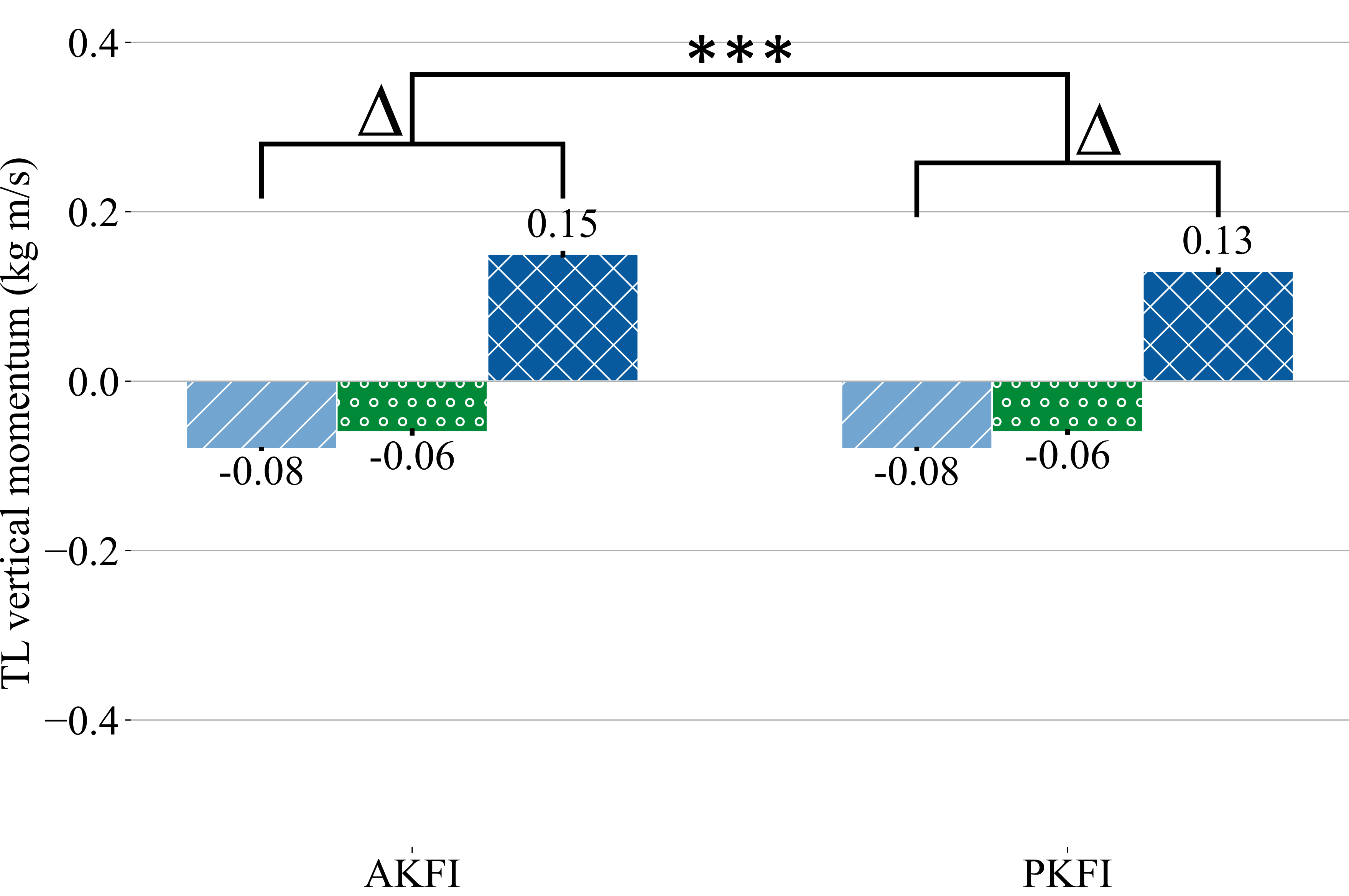}\label{fig:momentums:b}}
\hfil
\sidesubfloat[]{\includegraphics[width=0.48\textwidth]{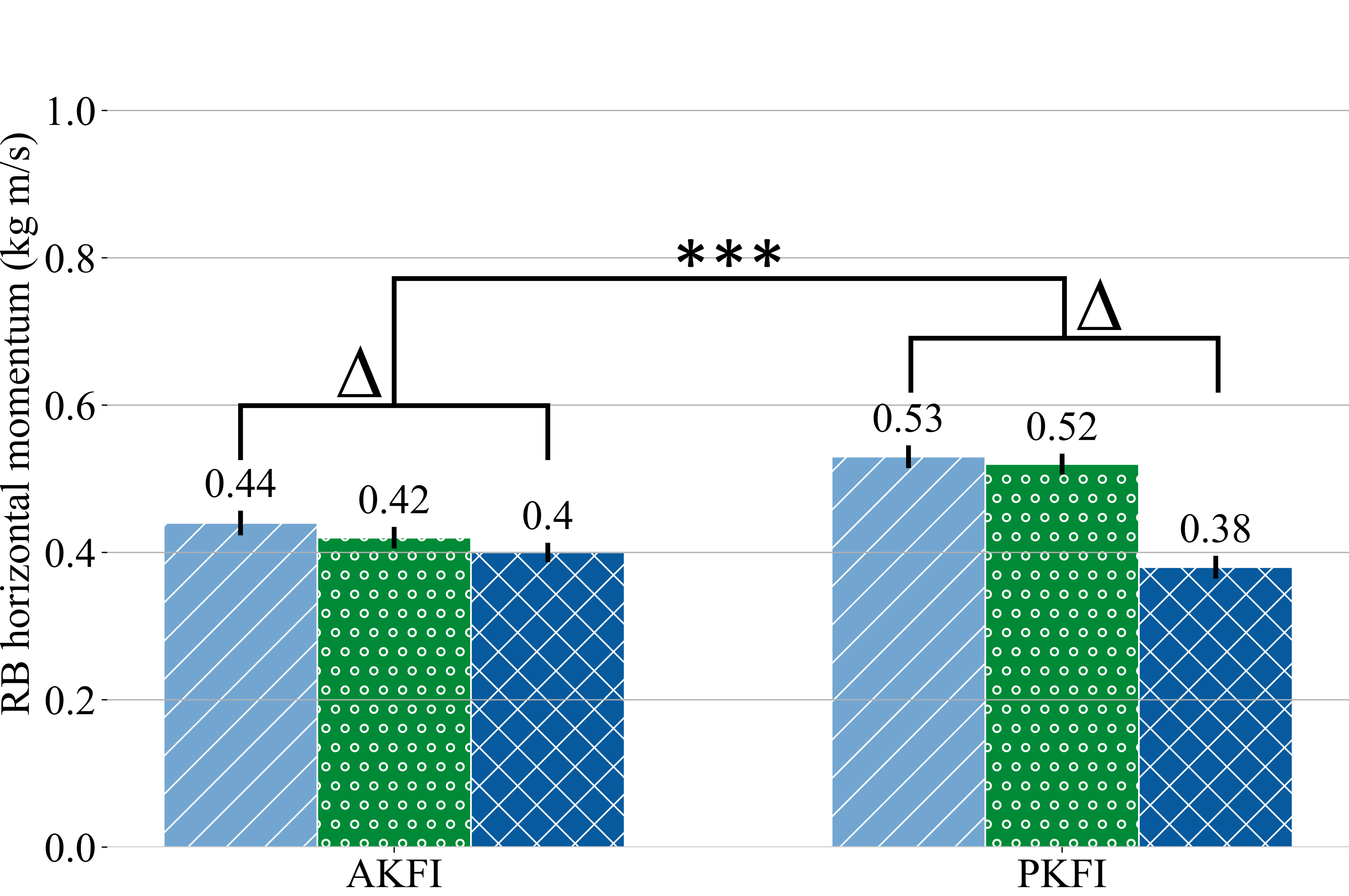}\label{fig:momentums:c}}
\hfil
\sidesubfloat[]{\includegraphics[width=0.48\textwidth]{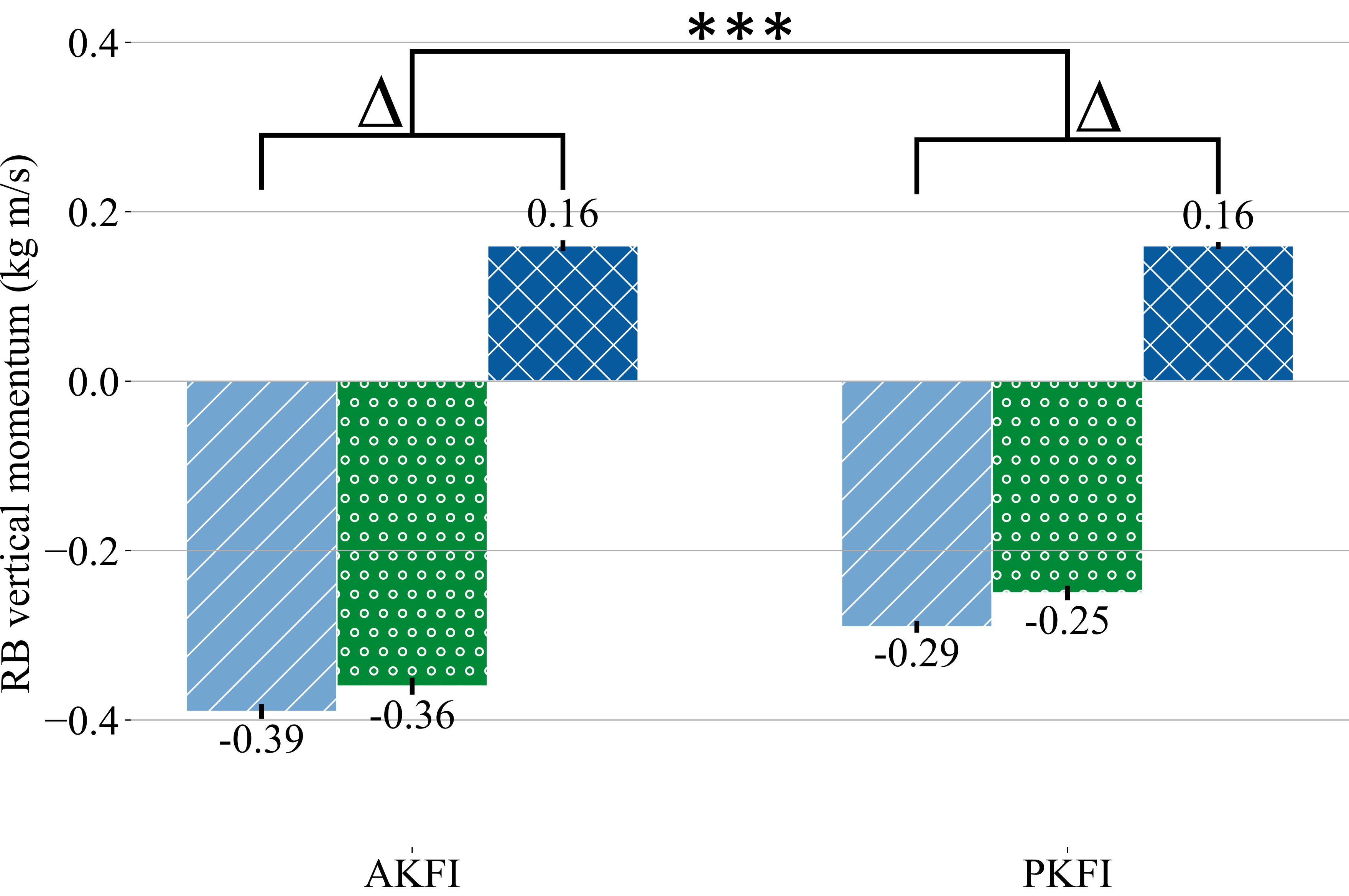}\label{fig:momentums:d}}
\hfil
\sidesubfloat[]{\includegraphics[width=0.48\textwidth]{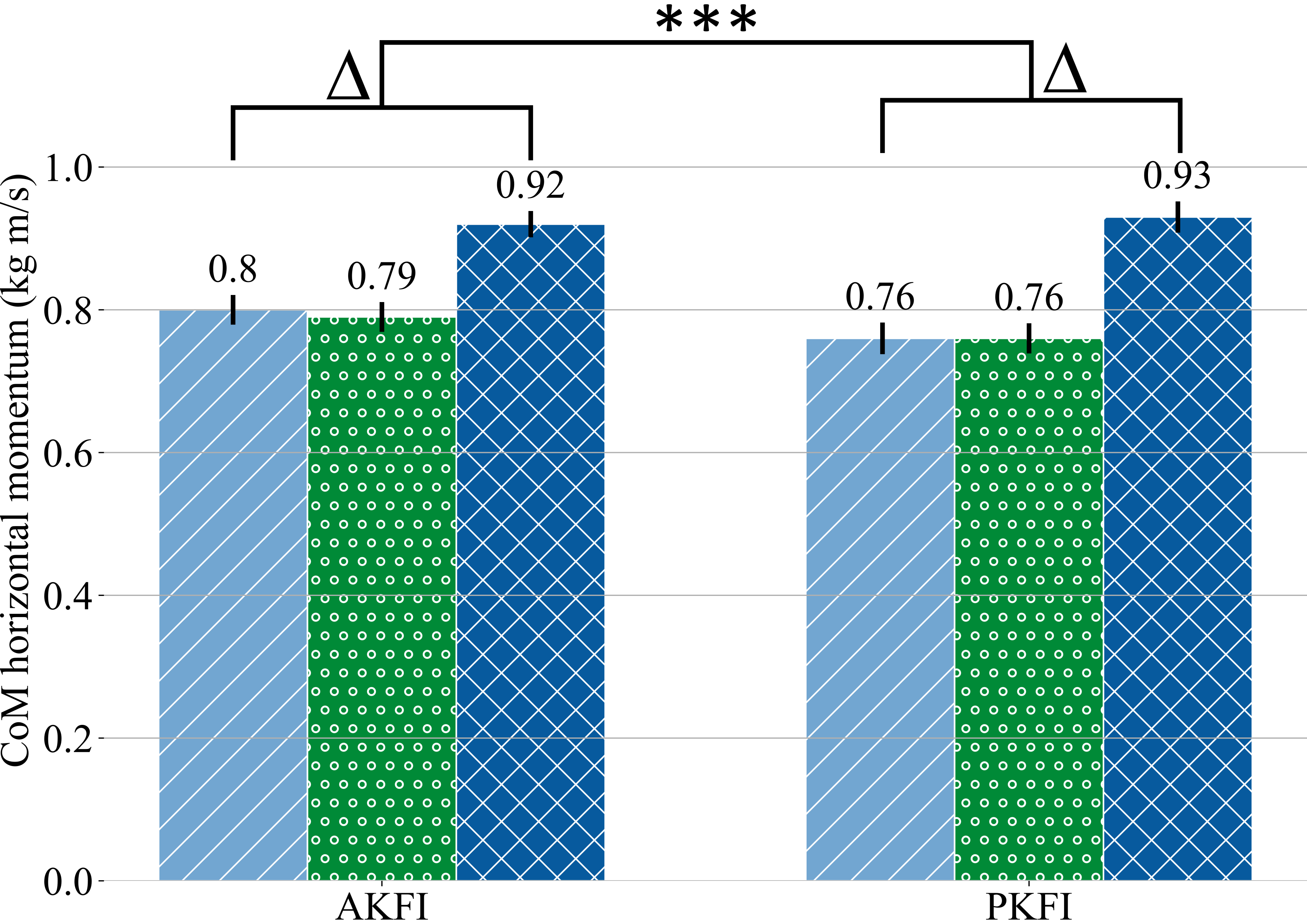}\label{fig:momentum_CoM:a}}
\hfil
\sidesubfloat[]{\includegraphics[width=0.48\textwidth]{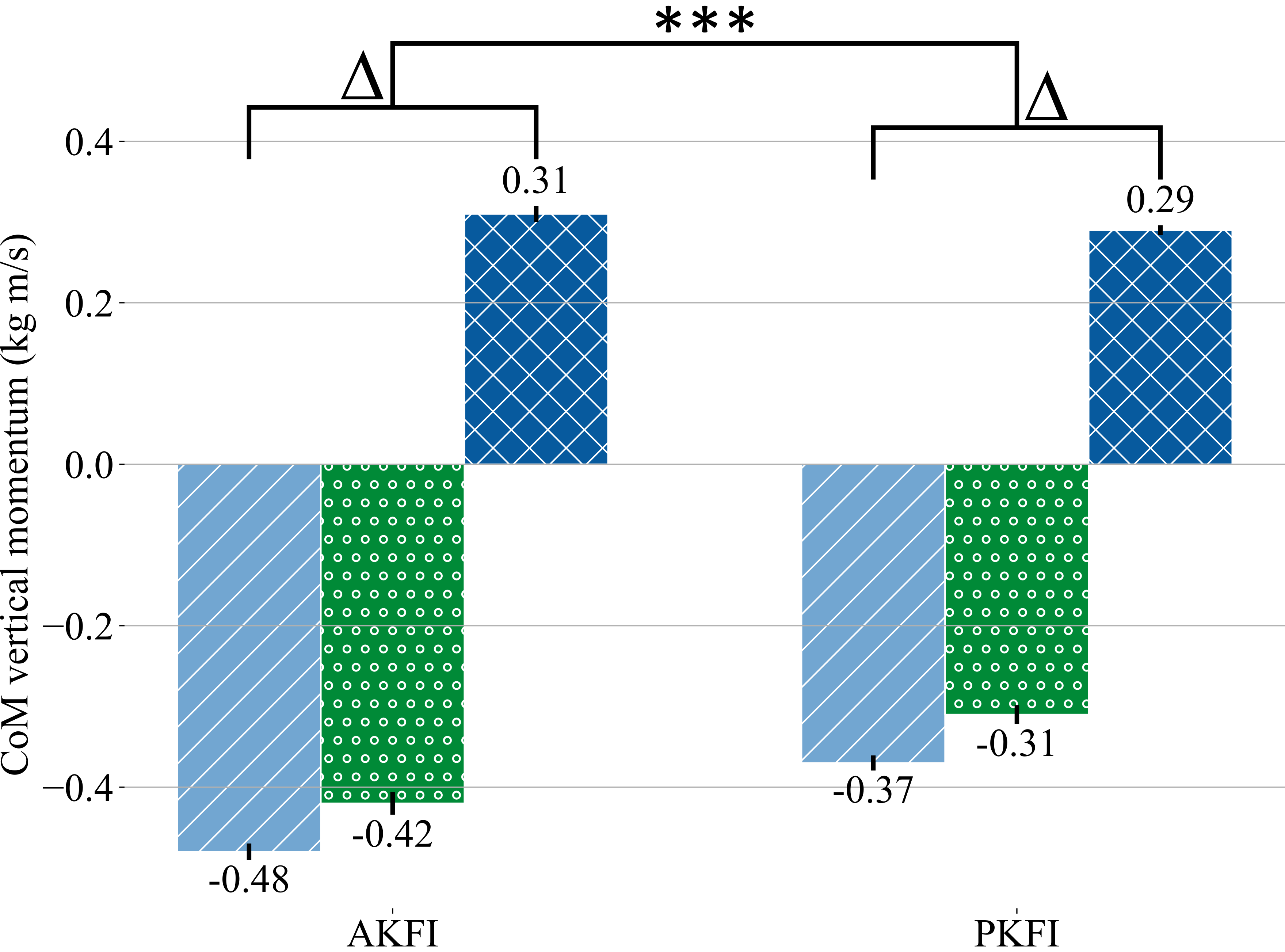}\label{fig:momentum_CoM:b}}
\caption{Trailing Leg (TL), Remaining Body (RB), and Center of Mass (CoM) instantaneous momentums in horizontal (\textbf{a}, \textbf{c}, and \textbf{e}) and vertical (\textbf{b}, \textbf{d}, and \textbf{f}) directions at the start of the step-to-step transition (\textcolor{sky_b}{vmin}), at leading leg touch-down (\textcolor{green}{LLTD}), and at the end of the step-to-step transition (\textcolor{sky_d}{vmax}) in the active knee flexion initiation (\akf{}) and in the passive knee flexion initiation (\pkf{}) experiments. The vertical black lines at the top of the bars show the standard deviations. The change in TL horizontal momentum is larger with \pkf{} than with \akf{}, and RB horizontal momentum decreases with \pkf{}. The CoM's vertical momentum increases more in \akf{} than in \pkf{} experiment during the step-to-step transition. The momentum values and their standard deviation values are available in Supplementary Table S3. *** denote significant difference between the momentum changes ($\Delta$) during the step-to-step transition period in the \akf{} and in the \pkf{} experiments with p < 0.001. The exact p values are available in \Cref{tab:statTest_Res}.}
\label{fig:momentums}
\end{figure*}

\begin{figure*}[ht]
\centering
\sidesubfloat[]{\includegraphics[width=0.48\textwidth]{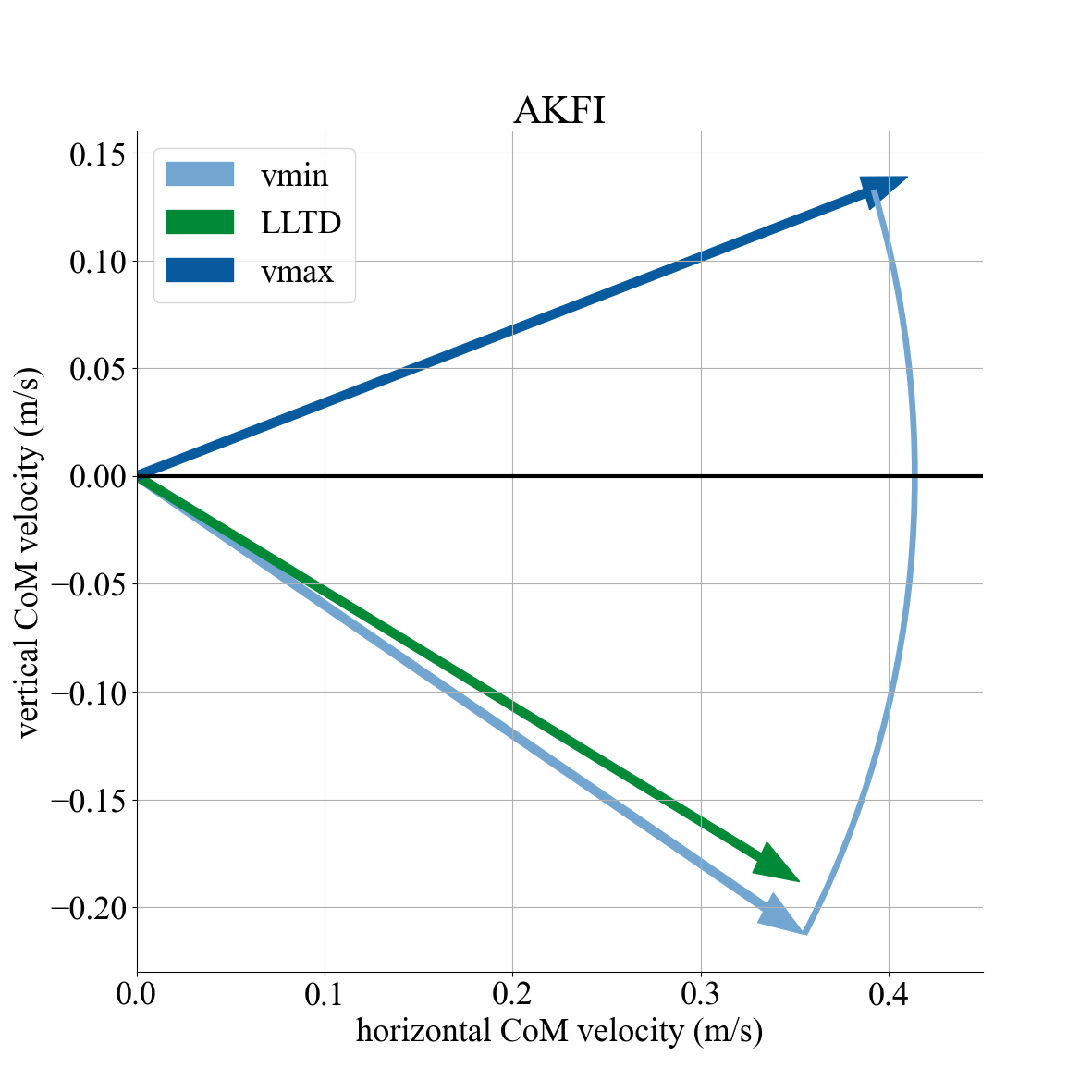}\label{fig:vel_vectors:a}}
\hfil
\sidesubfloat[]{\includegraphics[width=0.48\textwidth]{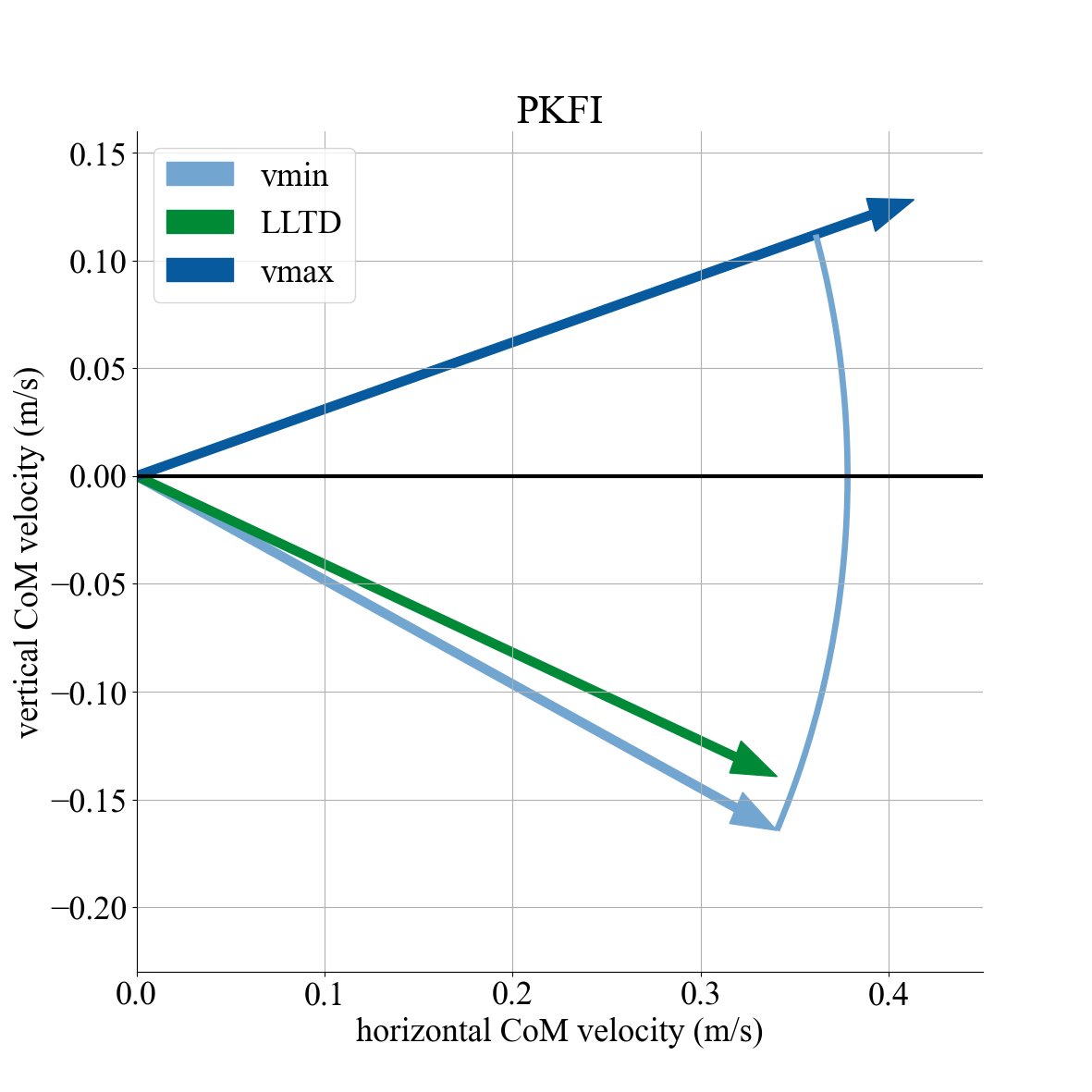}\label{fig:vel_vectors:b}}
\caption{Center of Mass (CoM) velocity vectors at the start of the step-to-step transition (\textcolor{sky_b}{vmin}), at leading leg touch-down (\textcolor{green}{LLTD}), and at the end of the step-to-step transition (\textcolor{sky_d}{vmax}) in the active knee flexion initiation (\akf{} - \textbf{a}) and in the passive knee flexion initiation (\pkf{} - \textbf{b}) experiments. An arc is drawn with a radius that is equal to the length of the velocity vector at vmin to better show the relation of the vector lengths at the three different times during the step-to-step transition. The length of the velocity vector increases between vmin and vmax more in the \pkf{} experiment than in the \akf{} experiment. }
\label{fig:vel_vectors}
\end{figure*}

Similar hip, knee, and ankle joint angles were reached with active knee flexion initiation (\akf{}) and with passive knee flexion initiation (\pkf{}) (\Cref{fig:joint_angles}). The robot could swing its leg with a flexed knee by passive knee flexion initiation (\Cref{fig:joint_angles}), allowing sufficient toe clearance during swing (see Supplementary Figures S3 and S4, and Supplementary Movies S1 and S2). The robot reached an average walking speed of \SI{0.44}{m/s} with \pkf{}, the same as the average walking speed with \akf{}. \\
We can observe a delay in the gait events in \pkf{} experiment compared to \akf{} experiment (\Cref{fig:timing_graph}). Leading leg touch-down (LLTD) happens at \SI{47}{\% gait cycle (GC)} and at \SI{48}{\% GC} with \pkf{} and with \akf{}, respectively. The knee joint starts to flex (Start of Knee Flexion - SKF) \SI{5}{\% GC} later, the hip joints start to flex (Start of Hip Flexion - SHF) \SI{4}{\% GC} later, and the ankle starts to plantarflex (Start of Ankle Plantar Flexion - SAPF) \SI{3}{\% GC} later with \pkf{} than with \akf{} (\Cref{fig:timing_graph}). The ankle starts to plantarflex (SAPF) \SI{2}{\% GC} before leading leg touch-down (LLTD) with \akf{} while the ankle plantarflexes (SAPF) only \SI{2}{\% GC} after leading leg touch-down (LLTD) with \pkf{} (\Cref{fig:timing_graph}). The time where the ankle starts to plantarflex (SAPF) indicates when the springs spanning the ankle joint start to unload with positive ankle joint power profile (see Supplementary Figure S9). \\
SKF occured significantly later (\SI{5}{\%GC}) in \pkf{} than in \akf{} experiment (p < 0.001) (\Cref{tab:statTest_Res}, \Cref{fig:timing_graph}, and Supplementary Table S2). The time period between SAPF and LLTD is significantly different with p < 0.001 between \akf{} and \pkf{} experiment (\Cref{tab:statTest_Res}, \Cref{fig:timing_graph}, and Supplementary Table S2). \\
The vertical and horizontal momentum changes of the trailing leg and remaining body are significantly different between the \akf{} and \pkf{} experiments \Cref{tab:statTest_Res}. The instantaneous momentum of the trailing leg increases by \SI{0.14}{kg m/s} more in the horizontal direction with \pkf{} compared to \akf{} during the step-to-step transition (\Cref{fig:momentums:a}, \Cref{tab:statTest_Res}, and Supplementary Figure S5). The trailing leg's momentum changes by \SI{0.23}{kg m/s} and by \SI{0.21}{kg m/s} in the vertical direction during the step-to-step transition in the \akf{} and \pkf{} experiments, respectively (\Cref{fig:momentums:b}, \Cref{tab:statTest_Res}, and Supplementary Figure S6). The horizontal momentum of the remaining body decreases by \SI{0.15}{kg m/s} with \pkf{} while it decreases by \SI{0.04}{kg m/s} with \akf{} during the step-to-step transition (\Cref{fig:momentums:c}, \Cref{tab:statTest_Res}, and Supplementary Figure S5). The remaining body's vertical momentum increases \SI{0.1}{kg m/s} less with \pkf{} than with \akf{} during the step-to-step transition (\Cref{fig:momentums:d}, \Cref{tab:statTest_Res}, and Supplementary Figure S6).  \\
Between the start of the step-to-step transition and leading leg touch-down, all trailing leg and remaining body momentum changes are within \SI{0.03}{kg m/s} in horizontal direction and within \SI{0.05}{kg m/s} in vertical direction in \pkf{} and in \akf{} experiments (\Cref{fig:momentums}, Supplementary Figure S5 and S6). More than \SI{50}{\%} of the above detailed changes happens between leading leg touch-down and the end of the step-to-step transition (\Cref{fig:momentums}, Supplementary Figure S5 and S6).\\
The vertical and horizontal momentum changes of the CoM are significantly different between the \akf{} and \pkf{} experiments \Cref{tab:statTest_Res}. The momentum of the CoM increases by \SI{0.16}{kg m/s} and by \SI{0.12}{kg m/s} in the horizontal direction during the step-to-step transition with \pkf{} and with \akf{}, respectively (\Cref{fig:momentum_CoM:a}, \Cref{tab:statTest_Res}, and Supplementary Figure S5). The CoM's vertical momentum increases by \SI{0.13}{kg m/s} more in \akf{} than in \pkf{} experiment during the step-to-step transition. At the end of the step-to-step transition, the vertical momentum of the CoM reaches \SI{0.31}{kg m/s} and \SI{0.29}{kg m/s} with \akf{} and with \pkf{}, respectively (\Cref{fig:momentum_CoM:b}, Supplementary Table S3, and Supplementary Figure S6). \\
The magnitude increase of the trailing leg momentum vector during the step-to-step transition is significantly larger (\SI{77}{\%}) in \pkf{} experiment than in \akf{} experiment (p < 0.001) (\Cref{fig:momentums}, \Cref{tab:statTest_Res}). The magnitude reduction of the remaining body momentum vector during the step-to-step transition is significantly larger (\SI{19}{\%}) in \pkf{} experiment than in \akf{} experiment (p < 0.001) (\Cref{fig:momentums}, \Cref{tab:statTest_Res}). The magnitude increase of the CoM momentum vector during the step-to-step transition is significantly larger (\SI{188}{\%}) in \pkf{} experiment than in \akf{} experiment (p < 0.001) (\Cref{fig:momentums}, \Cref{tab:statTest_Res}).\\
The direction change of the CoM velocity vector during the step-to-step transition is \SI{6.6}{deg} larger in the \akf{} experiment than in the \pkf{} experiment (\Cref{fig:vel_vectors}, and Supplementary Figure S7). The vertical component of the CoM velocity vector at the start of the step-to-step transition is \SI{0.05}{m/s} larger in the negative direction with \akf{} than with \pkf{} (\Cref{fig:vel_vectors}). At the end of the step-to-step transition, the horizontal components of the CoM velocity vectors are equal (\SI{0.41}{m/s}), and the vertical components are \SI{0.14}{m/s} and \SI{0.13}{m/s} in the \akf{} and in the \pkf{} experiment, respectively (\Cref{fig:vel_vectors}, and Supplementary Table S4). \\
The center of mass total kinetic energy at the start of the step-to-step transition is \SI{0.03}{J} lower in the \pkf{} experiment than in the \akf{} experiment. However, at the end of the step-to-step transition, the center of mass total kinetic energy reaches \SI{0.21}{J} in both \pkf{} and \akf{} experiments (see Supplementary Figure S8, and Supplementary Table S5). \\
The net positive cost of transport was \SI{0.56}{} (SD: \SI{0.05}{}) in the \akf{} experiment and \SI{0.52}{} (SD: \SI{0.03}{}) in the \pkf{} experiment. A natural runner's cost of transport with the same body mass {m=\SI{2.1}{kg}} is $\cotnr=\num{1.36}$ \cite{tucker_energetic_1970,birdbot_2022}. Hence, the relative cost of transport ($\cotre$) of the EcoWalker-2 robot is \SI{41}{\%} in the \akf{} experiment, and \SI{38}{\%} in the \pkf{} experiment, respectively.
%
%
%
%
%
%
%
%
%
\input{tables/table_statTest_Res}
%
%
%
%

%% file: tables/table_statTest_Res.tex
\begin{table*}[!ht]
\begin{tabularx}{0.98\textwidth}{X l | >{\centering\arraybackslash}X | >{\centering\arraybackslash}X >{\centering\arraybackslash}X | >{\centering\arraybackslash}X >{\centering\arraybackslash}X | >{\centering\arraybackslash}X}
\multicolumn{2}{c|}{\multirow{2}{*}{\textbf{measure}}} & \multicolumn{1}{c|}{\multirow{2}{*}{\textbf{p value}}} & \multicolumn{2}{c}{\textbf{AKFI}} & \multicolumn{2}{c|}{\textbf{PKFI}} &
  \multicolumn{1}{c}{\multirow{2}{*}{\textbf{diff. in \%}}} \\ \cline{4-7} 
\multicolumn{2}{c|}{}                                                             & \multicolumn{1}{c|}{} & \textbf{mean}  & \textbf{SD}   & \textbf{mean}  & \textbf{SD}   \\ \hline
\multirow{3}{*}{\textbf{gait event}} & $\mathrm{t}_\mathrm{SKF}$ (in \%GC)               & 1.90e-21             & 39.00 & 0.26 & 43.67 & 0.37 & 11.98                \\ \cline{2-8} 
                             & $\mathrm{t}_\mathrm{SAPF}$ (in   \%GC)            & 1.80e-21             & 46.25 & 0.30 & 48.77 & 0.18 & 5.45                 \\ \cline{2-8} 
                             & $\Delta\mathrm{t}_\mathrm{SAPF-LLTD}$   (in \%GC) & 1.90e-21             & 1.62  & 0.33 & -1.88 & 0.11 & -215.55              \\ \hline
\multirow{3}{*}{\textbf{TL imp.}}     & $\Delta|\textbf{p}_\mathrm{TL}|$ (in kg m/s)      & 1.97e-21             & 0.18  & 0.01 & 0.31  & 0.01 & 77.17                \\ \cline{2-8} 
                             & $\Delta\mathrm{p}_\mathrm{TL,x}$   (in kg m/s)    & 1.97e-21             & 0.17  & 0.01 & 0.31  & 0.01 & 87.31                \\ \cline{2-8} 
                             & $\Delta\mathrm{p}_\mathrm{TL,y}$   (in kg m/s)    & 1.97e-21             & 0.23  & 0.00 & 0.21  & 0.00 & -9.81                \\ \hline
\multirow{3}{*}{\textbf{RB imp.}}     & $\Delta|\textbf{p}_\mathrm{RB}|$ (in kg m/s)      & 1.60e-15             & -0.16 & 0.02 & -0.19 & 0.03 & 19.09                \\ \cline{2-8} 
                             & $\Delta\mathrm{p}_\mathrm{RB,x}$   (in kg m/s)    & 2.07e-21             & -0.04 & 0.03 & -0.15 & 0.03 & 245.83               \\ \cline{2-8} 
                             & $\Delta\mathrm{p}_\mathrm{RB,y}$   (in kg m/s)    & 1.97e-21             & 0.55  & 0.01 & 0.45  & 0.01 & -19.64               \\ \hline
\multirow{3}{*}{\textbf{CoM imp.}}    & $\Delta|\textbf{p}_\mathrm{CoM}|$ (in kg m/s)     & 4.74e-21             & 0.04  & 0.03 & 0.12  & 0.04 & 187.82               \\ \cline{2-8} 
                             & $\Delta\mathrm{p}_\mathrm{CoM,x}$   (in kg m/s)   & 2.30e-13             & 0.12  & 0.03 & 0.16  & 0.04 & 32.74                \\ \cline{2-8} 
                             & $\Delta\mathrm{p}_\mathrm{CoM,y}$   (in kg m/s)   & 1.97e-21             & 0.79  & 0.01 & 0.66  & 0.01 & -16.74
\end{tabularx}
\caption{Results of the Wilcoxon signed-rank tests (p values) to test the differences between the active knee flexion initiation (\akf{}) and the passive knee flexion initiation (\pkf{}) experiments. The tested measures were: time of the start of knee flexion ($\mathrm{t}_\mathrm{SKF}$), time of the start of ankle plantar flexion ($\mathrm{t}_\mathrm{SAPF}$), the time period between the start of ankle plantar flexion and the leading leg touch-down ($\Delta\mathrm{t}_\mathrm{SAPF-LLTD}$), the absolute ($\Delta|\textbf{p}_\mathrm{TL}|$), horizontal ($\Delta\mathrm{p}_\mathrm{TL,x}$), and vertical ($\Delta\mathrm{p}_\mathrm{TL,y}$) momentum change (impulse) of the trailing leg during the step-to-step transition, the the absolute ($\Delta|\textbf{p}_\mathrm{RB}|$), horizontal ($\Delta\mathrm{p}_\mathrm{RB,x}$), and vertical ($\Delta\mathrm{p}_\mathrm{RB,y}$) momentum change (impulse) of the remaining body during the step-to-step transition, and the the absolute ($\Delta|\textbf{p}_\mathrm{CoM}|$), horizontal ($\Delta\mathrm{p}_\mathrm{CoM,x}$), and vertical ($\Delta\mathrm{p}_\mathrm{CoM,y}$) momentum change (impulse) of the center of mass during the step-to-step transition. All tested differences were significant with a significance level of p < 0.001. \textbf{Abbreviations:} x: horizontal direction, y: vertical direction, TL: Trailing Leg, RB: Remaining Body, CoM: Center of Mass of the whole robot, imp.: impulse = change in momentum, SD: standard deviation, diff. in \% was calculated as: $\frac{\mathrm{mean}_\mathrm{PKFI}- \mathrm{mean}_\mathrm{AKFI}}{\mathrm{mean}_\mathrm{AKFI}} \cdot 100$.}
\label{tab:statTest_Res}
\end{table*}

%% file: Sections/Discussion.tex
\section{Discussion}
%
%
%
The first goal of this study was to test Perry's \cite{Perry1992} passive knee flexion process on a bipedal robot, as part of the lock-and-release mechanism of the human swing leg catapult. We showed that the EcoWalker-2 robot is able to walk with sufficient toe clearance both with passive knee flexion initiation (\pkf{}) and with active knee flexion initiation (\akf{}).
The second aim of the study was to influence the ratio of energy flow into the remaining body and the trailing leg by indirectly manipulating the timing of the start of the ankle plantar flexion. We showed that ankle plantar flexion starts significantly later in the \pkf{} experiment than in the \akf{} experiment. This delay had a pronounced effect on the momentum changes during the step-to-step transition. Notably, the magnitude increase in the trailing leg momentum vector during the step-to-step transition was significantly larger (\SI{77}{\%}), the magnitude decrease in the remaining body momentum vector during the step-to-step transition was significantly larger (\SI{19}{\%}), and the magnitude increase in the center of mass momentum vector during the step-to-step transition was significantly larger (\SI{188}{\%}) in the \pkf{} experiment than in the \akf{} experiment. \\
%
%
%
As expected, starting the knee flexion earlier in the \akf{} experiment enables an earlier onset of ankle plantar flexion (\Cref{fig:timing_graph}). The timing delay of the start of ankle plantar flexion gives further evidence of Perry's hypothesis that knee flexion is needed for ankle unloading \cite{Perry1992}.
Surprisingly, leading leg touch-down did not happen exactly at \SI{50}{\% of the gait cycle (GC)} in either \pkf{} or \akf{} experiment by neither legs (\Cref{fig:timing_graph}, see Supplementary Information S2). The 1-\SI{3}{\%GC} earlier or later leading leg touch-down could be a result of the slight asymmetry between the right and left sides of the robot. Even though the slack positions of the GAS and SOL spring-tendons were set symmetrically, there was always a little difference between the left and right, hip and knee joint actuators which causes a minor asymmetry in the robot gait. Asymmetry in human gait is also a common phenomenon, with strength imbalances present even in young adults. Research has shown that young adults typically exhibit strength asymmetries ranging from \SI{5}{\%} to \SI{15}{\%} \cite{Perry2007, Lanshammar2011}. In contrast, older adults tend to develop more pronounced asymmetries, typically in the range of \SI{15}{\%} to \SI{20}{\%} \cite{Skelton2002, Perry2007}. 

The time period between the start of ankle plantar flexion and leading leg touch-down was expected to be shorter in \pkf{} experiment than in \akf{} experiment. However, the delay in ankle plantar flexion onset was more pronounced than expected in the \pkf{} experiment. Specifically, ankle plantar flexion occurred \SI{2}{\%GC} before leading leg touch-down in the \akf{} experiment, similar to human data \cite{VanderZee2022}, whereas it occurred \SI{2}{\%GC} after leading leg touch-down in the \pkf{} experiment. The start of ankle plantar flexion means the start of ankle energy unloading where the ankle power curve becomes positive. Both in the \pkf{} and \akf{} experiments, the event flow of the step-to-step transition can only start after the start of knee flexion. As the start of knee flexion is delayed in the \pkf{} experiment, all gait events of the trailing leg are delayed. In human experimental gait data, the start of knee flexion was reported to happen before the start of ankle plantar flexion and before leading leg touch-down \cite{Perry1992, Lipfert2014}. Knee flexion appears to play a crucial role in initiating the step-to-step transition, with its timing having a cascading effect on subsequent gait events. \\
%
%
%
With the \pkf{} and \akf{} experiments, we characterized two distinct disengagement processes on the same robotic system. \\
In the \akf{} experiment, as ankle unloading started before leading leg touch-down, hip and knee flexion pulled the ankle joint upwards, making it possible for the ankle joint to start plantar flexing. Part of the unloading happened against the whole body weight before leading leg touch-down, a mechanism known as preemptive push-off, which has been well documented in human experiments \cite{Fukunaga2001, ishikawa2005muscle, Lichtwark.2007, Cronin2013, Perry1992}. In humans, preemptive push-off is initiated by gastrocnemius (GAS) and soleus (SOL) muscle activations, while in the robot, it is initiated by active hip and active knee flexion. After leading leg touch-down, only elastic recoil happens in humans as the observed GAS and SOL muscle activations diminish \cite{Fukunaga2001, ishikawa2005muscle, Lichtwark.2007, Cronin2013, Perry1992}. After leading leg touch-down, the robot's hip and knee joints remain active, facilitating the rest of ankle unloading against the trailing leg through a dual mechanism of elastic recoil and active knee and hip flexion. \\
In literature, this preemptive push-off is explained as a method to restore heel-strike collision losses \cite{McGeer.1990, Donelan2002_mechWork, Donelan2002_simPosNeg, Kuo2002, Collins2005, Dean2009}. By using active knee and hip flexion with a passive ankle joint, we could reach a similar preemptive push-off. Notably, increasing hip work to facilitate leg swing has been shown to reduce ankle muscle activations during push-off \cite{Lenzi2013}, which can be beneficial for reducing peak plantar pressures and treating conditions such as neuropathic plantar ulcers \cite{Mueller1994}. Conversely, studies have found that increasing ankle push-off work can decrease hip work \cite{Lewis2008, Caputo2014, Koller2015}, highlighting the trade-offs between hip and ankle contributions during gait. \\
In the \pkf{} experiment, ankle unloading started only after leading leg touch-down, because hip flexion without active knee flexion initiation did not result in an upward pull at the ankle joint. As the ankle joint is passive, only elastic recoil from the GAS and SOL spring-tendons actuated the ankle plantar flexion. Unloading happened mainly against the trailing leg, as leading leg touch-down already happened. \\
%
%
%
Before leading leg touch-down, the trailing leg bears the whole body weight as the leading leg is in swing. We hypothesized that the ankle energy being unloaded before leading leg touch-down mainly accelerates the remaining body. As the leading leg takes up increasingly more body weight after leading leg touch-down, the trailing leg becomes increasingly free to go into swing. We expected that the proportion of ankle energy unloaded after leading leg touch-down would determine the distribution of energy between the trailing leg and the remaining body. A greater unloading of ankle energy after leading leg touch-down would result in a larger transfer of energy to the trailing leg, accelerating it into swing, while a smaller amount of energy would accelerate the remaining body. In the \akf{} experiment, a small part of the ankle energy was already unloaded before leading leg touch-down but no part of the ankle energy was unloaded before leading leg touch-down in the \pkf{} experiment (see Supplementary Figure S9). To check our hypothesis, we looked at the magnitude of the trailing leg and remaining body momentum vectors, and their horizontal and vertical components (\Cref{tab:statTest_Res}, \Cref{fig:momentums}) at the start of the step-to-step transition (vmin), at leading leg touch-down, and at the end of the step-to-step transition (vmax). \\
%
%
%
In line with our hypothesis, the trailing leg momentum vector's magnitude increased significantly more in the \pkf{} experiment than in the \akf{} experiment during the step-to-step transition (\Cref{tab:statTest_Res}). The direction of the trailing leg momentum vectors at the start and end of the step-to-step transition reveals that the significant difference in magnitude increase is primarily due to the significant difference in the increase of the horizontal component of the trailing leg momentum. (\Cref{fig:momentums}). The trailing leg horizontal momentum started to increase at \SI{\sim 20}{\%GC} in both \akf{} and \pkf{} experiments, but the increasing trend stopped when the knee motor torque command turned zero in the \pkf{} experiment (see Supplementary Figure S5). The trailing leg horizontal momentum started to increase again only after the start of knee flexion in the \pkf{} experiment. Consequently, the trailing leg horizontal momentum reached a higher value until the start of the step-to-step transition in the \akf{} experiment than in the \pkf{} experiment (see Supplementary Figure S5). Nevertheless, the trailing leg horizontal momentum increased at a higher rate in the \pkf{} experiment than in the \akf{} experiment, reaching a similar value at the end of the step-to-step transition. Similarly, human experimental walking data at 50-\SI{100}{\%} of the preferred transition speeds show a positive horizontal, vertical, and absolute trailing leg momentum change between the start of ankle plantar flexion and toe-off \cite{Lipfert2014}. The similarity between our experimental data and human walking data supports the idea that the observed trailing leg momentum dynamics are a key feature of efficient human gait.\\
%
%
Validating our hypothesis, the remaining body momentum increased less in the \pkf{} experiment than in the \akf{} experiment. Actually, the magnitude decrease of the remaining body momentum vector during the step-to-step transition was significantly larger in the \pkf{} experiment than in the \akf{} experiment (\Cref{tab:statTest_Res}). The remaining body momentum vector change during the step-to-step transition was different between the \pkf{} and the \akf{} experiment in both horizontal and vertical directions  (\Cref{fig:momentums}). The horizontal component of the remaining body momentum vector decreased more during the step-to-step transition in the \pkf{} experiment than in the \akf{} experiment, but it reached similar momentum values of \SI{0.4}{kg m/s} and \SI{0.38}{kg m/s} in the \akf{} and in the \pkf{} experiment, respectively. The vertical component of the remaining body momentum vector increased more during the step-to-step transition in the \akf{} experiment than in the \pkf{} experiment, but it reached the same \SI{0.16}{kg m/s} value at the end of the step-to-step transition. The remaining body horizontal momentum started to decrease at \SI{\sim 35}{\%GC} (see Supplementary Figure S5), while the remaining body vertical momentum started to decrease at \SI{\sim 30}{\%GC} (see Supplementary Figure S6) in both \akf{} and \pkf{} experiment, but the decrease rates were larger in the \akf{} experiment than in the \pkf{} experiment. By the start of the step-to-step transition, the remaining body horizontal and vertical momentums reached a lower value in the \akf{} than in the \pkf{} experiment (see Supplementary Figure S5, see Supplementary Figure S6). The larger rates of horizontal and vertical momentum decrease before the start of the step-to-step transition could be a consequence of the earlier start of knee flexion in the \akf{} experiment. Similarly, human experimental walking data at 50-\SI{100}{\%} of the preferred transition speeds show a negative horizontal, and positive vertical and absolute remaining body momentum change between the start of ankle plantar flexion and toe-off \cite{Lipfert2014}. These results suggest that our experiments effectively capture the essential dynamics of human walking, particularly with respect to the remaining body momentum changes. \\
%
%
The trailing leg and remaining body horizontal momentums were complementing each other in both \akf{} and \pkf{} experiments, which makes the center of mass (CoM) horizontal momentum similar at vmin, leading leg touch-down, and vmax in the \akf{} and \pkf{} experiments (\Cref{fig:momentum_CoM:a}, see Supplementary Figure S5). As the trailing leg vertical momentum change was similar in the \akf{} and \pkf{} experiments, the vertical momentum change of the remaining body is showing in the CoM vertical momentum changes (\Cref{fig:momentum_CoM:b}, see Supplementary Figure S6). The vertical momentum change difference comes from the vertical velocity component change of the CoM during the step-to-step transition (\Cref{fig:vel_vectors}). As the CoM horizontal velocity components were similar in the \akf{} and \pkf{} experiments, the change in the CoM vertical velocity components show as the change in the CoM velocity vector angles during the step-to-step transition (see Supplementary Figure S7). Even though the CoM velocity vector angle change was larger in the \akf{} experiment than in the \pkf{} experiment, the magnitude of the CoM velocity vector increased more in the \pkf{} experiment than in the \akf{} experiment (\Cref{fig:vel_vectors}). The changes in the CoM velocity vector magnitudes show in the CoM total kinetic energy changes (see Supplementary Figure S8). As a result of a larger negative remaining body vertical momentum at the start of the step-to-step transition, the total kinetic energy of the CoM was larger at the start of the step-to-step transition in the \akf{} than in the \pkf{} experiment. As the remaining body vertical momentums were equal and the trailing leg vertical momentums were similar, the total kinetic energies of the CoM were equal at the end of the step to step transition in the \akf{} and \pkf{} experiments. \\
%
%
%
While comparing robots with varying morphologies and numbers of legs has its limitations, it still offers valuable insights into how design of mechanics, actuators, and control influence energetic performance. Notably, the EcoWalker-2 robot's best relative net cost of transport of \SI{38}{\%}) is comparable to that of BiartHopper's \cite{ruppert_series_2019}, which achieved a relative net cost of transport of \SI{45}{\%}. However, the latter robot features only a single hip actuator, whereas EcoWalker-2 robot utilizes two knee and two hip actuators for planar walking.
With its design inspired by passive dynamic walkers, the Cornell biped \cite{Collins2005} achieved a relative cost of transport between \SI{7}{\%} and \SI{26}{\%} of a natural runner of similar weight. The calculation was based on the provided\cite{Collins2005} cost of transport range from 0.05 (mechanical) to 0.2 (total, electrical) at \SI{13}{kg} body weight. Therefore, EcoWalker-2 robot approaches the efficiency of passive-dynamic walkers, while remaining fully controllable and versatile. We identify three causes for the EcoWalker-2 robot's outstanding energetic efficiency. First, its brushless motors, low gear ratio, and efficiently built actuators contribute to the robot's low cost of transport \cite{Grimminger2019}. Second, the EcoWalker-2 robot walks with straight legs in mid-stance, i.e., when ground reaction forces are the highest. Without an acting moment arm, the robot's knee actuator is effectively unloaded around midstance. Third, the EcoWalker-2 robot's leg design and control allows harnessing the elastic energy in the spring-loaded ankles. The robot's catapult mechanism then converts the stored energy into combined swing leg motion and forward momentum. So far, the EcoWalker-2 robot is freely walking in the sagittal plane, but it is constrained against lateral motions and pitch rotation. Once a 3D-walking robot, we expect EcoWalker-2 robot's cost of transport to increase due to the required, additional actuation. \\
%
%
%
Our study has limitations as the experiments were done on a bipedal robot which can only walk in sagittal plane and its upper body is stabilized by a 4-bar mechanism. The restrictions of the robot's design and experimental setup could have potentially over emphasized the results as the upper body stabilization and 3D stabilization of the robot were not factored in. In our opinion, concentrating on the legs in the robot's design and on a single plane made it possible to evaluate the sagittal plane effects more clearly with less confounding factors. \\
Human gait data is highly variable, especially if it is about patient population \cite{VanCriekinge2023}, whereas gait events in robots are more reliable, SD < \SI{0.4}{\%GC} (see Supplementary Table S2, and Supplementary Table S6).
We found that the time sequence of gait events around the step-to-step transition (leading leg touch-down, start of ankle plantarflexion, start of knee and hip flexion) from human gait data is not defined clearly in literature \cite{Perry1992}\cite{Whittle2007}\cite{Richie2020}. A robotic system makes the gait event identification more reliable, which in turn makes gait analysis, based on gait events, and analysis of event flows more accurate. On a robotic system, we can investigate the push-off event flow, momentum and energy flow more reliably than with human measurements.
However the transfer of the results to human gait remains to be explored in future studies. \\
%
%
%
Changing the way of knee flexion initiation changes the timing of the gait events, remaining body, trailing leg and CoM velocities, momentums, and kinetic energies during the step-to-step transition. Although the method of knee flexion initiation has a significant impact during the step-to-step transition, its effect diminishes thereafter and remains negligible for the first \SI{\sim 20}{\%} of the gait cycle. Moreover, in terms of walking speed and cost of transport, the way of knee flexion initiation does not have a large overall effect on the whole gait cycle which also shows the robustness of human-like walking gaits. Changes in the gait during the step-to-step transition change the interplay between the disengagement process of the trailing leg and the collision process of the leading leg with the ground. \\
By manipulating the way of knee flexion initiation, we showed that the timing of the gait events around push-off, the distribution of kinetic energy, showing in velocity and momentums of the trailing leg and remaining body, the redirection of the CoM velocity vector and the CoM kinetic energy can be influenced. Knee flexion initiation has been identified as a critical factor in ankle unloading and the coordination of gait events at the end of stance, providing insights into the biomechanical mechanisms underlying the step-to-step transition in human walking. These findings have implications for the development of assistive technologies, including orthotics and prosthetics, and may inform gait rehabilitation practices. By optimizing knee flexion timing in these devices and protocols, clinicians and engineers may be able to restore more natural leg dynamics and improve walking efficiency in individuals with gait impairments.

%% file: Sections/Methods.tex
\section{Methods}
\begin{figure*}[ht]
    \centering
    \includegraphics[width=0.7\linewidth]{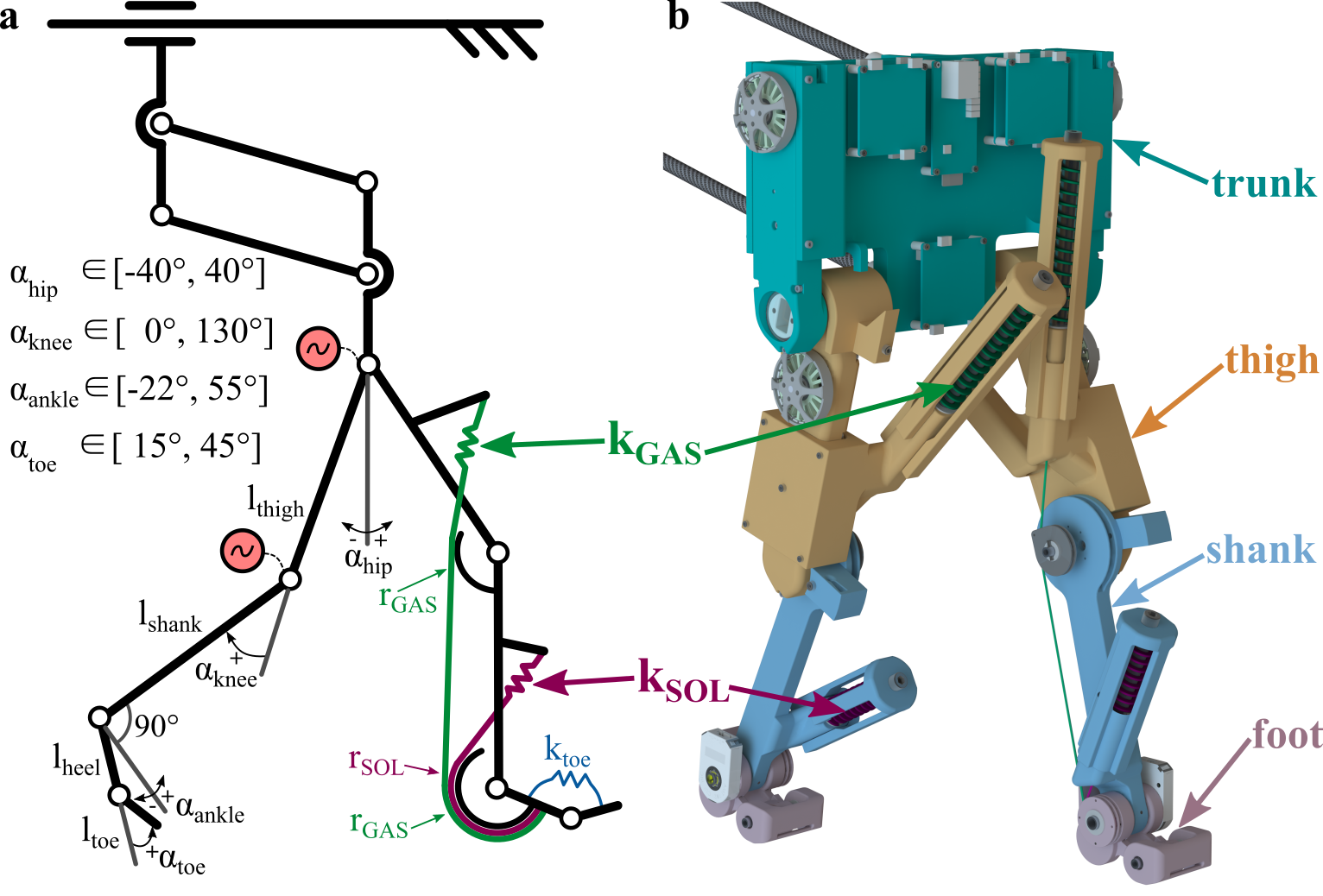}
    \caption{Schematic and rendering of the bipedal EcoWalker-2 robot. \textbf{a:} Schematic of the robot with spring-tendon routing, angle definitions, and range of motion of the joints. Segment lengths: $\mathrm{l}_\mathrm{thigh}=\mathrm{l}_\mathrm{shank}=\SI{160}{mm}$, $\mathrm{l}_\mathrm{heel}=\SI{32}{mm}$, $\mathrm{l}_\mathrm{toe}=\SI{17}{mm}$. Pulley radii: $\mathrm{r}_\mathrm{GAS}=\mathrm{r}_\mathrm{SOL}=\SI{13}{mm}$. Spring stiffness values: $\mathrm{k}_\mathrm{GAS}=\SI{1.4}{N/mm}$, $\mathrm{k}_\mathrm{SOL}=\SI{4.5}{N/mm}$, $\mathrm{k}_\mathrm{toe}=\SI{8.04}{Nmm/deg}$. The robot's trunk movement is constrained to translations in the sagittal plane, and all trunk rotations are locked. The four-bar guide slides freely in fore-aft direction. \textbf{b:} Rendering of the bipedal robot with segment name indications.}
    \label{fig:robot_design}
\end{figure*}
The EcoWalker-2 robot, a bipedal, human-like robot with hip and knee motors, and spring-loaded passive ankles, was used in the experiments. Building upon the original EcoWalker design \cite{Kiss2022}, the updated EcoWalker-2 received an update to its design freeing up the movement of the knee joint while preserving toe clearance during swing phase. The robot's control was updated to switch the knee actuation passive during specified times of the gait cycle (GC). A calibration step was also introduced prior to each experiment to symmetrically set the slack length of the ankle spring-tendons. We conducted experiments with the EcoWalker-2 robot with active knee joint actuation during the whole gait cycle (\akf{}) and with passive knee joint actuation in more than \SI{50}{\% GC} (\pkf{}). To ensure comparability, the hip control, ankle spring-tendon properties, and slack lengths were kept identical in both experiments.
\subsection{Update of the bipedal robot design}
To enable passive knee flexion in the EcoWalker-2 robot, the passive knee extending structures of the EcoWalker-1 \cite{Kiss2022} were removed. The toe tendon, which pulled up the toes for ground clearance when the knee was flexed, and the VAS spring-tendon, representing the vasti muscle-tendon unit, were removed. In the EcoWalker-2 robot, only the GAS spring-tendon, representing the gastrocnemius muscle-tendon unit, and a SOLO actuator module \cite{Grimminger2019} act at the knee joint. The SOLO actuator modules are driven by brushless DC (BLDC) motors. To maintain toe clearance during swing, the robot's heel segment was shortened from \SI{48}{mm} to \SI{32}{mm} and the robot's toe segment was shortened from \SI{24}{mm} to \SI{17}{mm} (\Cref{fig:robot_design}). To dampen the touch-down of the heel, a \SI{2}{mm} thick silicone layer (shore hardness 00-50) was glued to the bottom of the heel segment. The ankle joints are passive, spring-loaded by the SOL and GAS spring-tendons, representing the soleus and gastrocnemius muscle tendon units. Prior to each experiment, a calibration step was introduced to set the slack length of the SOL and GAS ankle spring-tendons symmetrically and accurately. To maintain constant slack lengths of the spring-tendons throughout the experiments, we employed two measures. Firstly, anti-vibration glue (Vibra-Tite VC-3 Threadmate) was applied to the setscrews used for adjusting the slack lengths, thereby securing them in place. Secondly, a spring cage was installed around the SOL and GAS spring mechanisms to prevent any unwanted rotation. \\
In the EcoWalker-2 robot, no changes were made to the hip, and toe joint actuators of the original EcoWalker robot \cite{Kiss2022}. The hips are actuated by SOLO actuator modules, and rotational toe springs act around the toe joints, representing the foot compliance (\Cref{fig:robot_design}). \\
The stiffnesses of the used springs are the following: SOL spring: \SI{4.5}{N/mm}, GAS spring: \SI{1.4}{N/mm}, toe spring: \SI{8.04}{Nmm/deg}. The SOL and GAS spring-tendons are acting on the ankle at a constant moment arm of \SI{13}{mm}, and the GAS spring-tendon is acting on the knee joint at a constant moment arm of \SI{13}{mm}. The slack length of both the SOL and GAS spring-tendons were set to \SI{0}{mm} at \SI{-22}{\degree} ankle joint and \SI{0}{\degree} knee joint angles for both legs in all experiments. The rotational toe springs are loaded in stance from an initial toe angle of \SI{15}{\degree} where the toe spring torque is zero. Nomenclature and the zero-angle definitions are shown in \Cref{fig:robot_design}.\\
\subsection{Update of the robot control}
Central Pattern Generators (CPGs), amplitude controlled phase oscillators, were implemented for both hip and knee joints with a half-cycle phase shift between left and right oscillator nodes \cite{ijspeert2007swimming, Kiss2022}. Reference trajectories for both joints were extracted from a 2D neuromuscular simulation of human walking \cite{Geyer.2010} to design the commanded joint trajectories in the CPG control. The same CPG control parameters were used in all experiments (\Cref{tab:cpg_pm}) to make them comparable. The frequency ($f$) defines the length of one gait cycle, flexion factors ($F_\mathrm{hip}$ and $F_\mathrm{knee}$) define the start of hip and knee flexion separately, as a fraction of the gait cycle. The knee amplitude ($\Theta_\mathrm{kneeAmplitude}$) and offset ($\Theta_\mathrm{kneeOffset}$) parameters define the knee trajectory. The hip amplitude ($\Theta_\mathrm{hipAmplitude}$) and offset ($\Theta_\mathrm{hipOffset}$) parameters define the hip trajectory. The hip swing steady parameter ($\varphi_\mathrm{hip}$) defines the hip's steady duration at the end of the swing, as a fraction of a gait cycle. \\
\input{tables/table_CPG_pm}
The CPG-generated hip and knee reference trajectories were the input of a PD controller which calculated the commanded hip and knee motor currents. Due to slight mechanical differences, the hip actuators of the right and left leg performed mildly different. Hence, P gains of the hip motors were manually tuned to obtain matching left and right leg kinematics: $k_{p,\mathrm{hip, right}} = 30$, $k_{p,\mathrm{hip, left}} = 26$. The D gain of the hips, and the PD gains of the knees were set symmetrically: $k_{d,\mathrm{hip, right/left}} = 0.2$, $k_{p,\mathrm{knee, right/left}} = 7$, $k_{d,\mathrm{knee, right/left}} = 0.2$.\\
In the active knee flexion initiation (\akf{}) control set the knee motor was controlled to follow the CPG-defined knee angle curve in the whole gait cycle. For the passive knee flexion initiation (\pkf{}) control set the knee motor's control was modified to send zero torque command to the motor in the specified time window during a gait cycle (see Supplementary Figure S2). The \pkf{} mode was activated through a keyboard input, once the robot was in steady-state locomotion, typically after 2 gait cycles. \\ 
\subsection{Experimental setup and data acquisition}
We compared two knee motor modes: \pkf{} and \akf{}.
In the \pkf{} experiment, the start of the zero knee motor torque period was set to \SI{35}{\% GC} (see Supplementary Figure S2). The start of the period was chosen in the middle of the stance phase of the robot, so that the start time is after the loading response but before the start of push-off. The start of push-off was defined as the start of ankle plantar flexion. The knee motor torque remained zero until the end of the same gait cycle (see Supplementary Figure S2). One gait cycle starts at the touch-down of the foot and lasts until the next touch-down of the same foot. \\
To ensure the independence of our results from the exact start time of the zero knee motor torque period, we conducted an additional experiment where the zero torque was applied to the knee motor at \SI{40}{\% GC}, instead of \SI{35}{\% GC} (PKFI40 experiment). The data analysis of the PKFI40 experiment is available in the Supplementary Information S3. \\
In the \pkf{} experiment, knee flexion onset emerged dynamically. We observed a \pkf{} knee flexion onset at \SI{47}{\% GC} (\Cref{fig:timing_graph}). The CPG-controlled active knee flexion initiation in the \akf{} experiment was set prior to \SI{47}{\% GC}, at \SI{40}{\% GC}. Due to a \SI{\sim 20}{\% GC} shift between the observed touch-down of the foot and the start of the CPG defined gait cycle, \SI{40}{\% GC} translates into $F_\mathrm{knee}=0.6$ knee flexion factor in the CPG parameters (\Cref{tab:cpg_pm}). \\
Hip motor control and the CPG parameters (\Cref{tab:cpg_pm}) were not changed between the \pkf{} and \akf{} experiments. The robot’s trunk was constrained to translations in the sagittal plane, and all trunk rotations were locked with a four-bar mechanism in all experiments.\\
The setup was instrumented to measure the kinematics of the robot with rotary encoders at the hip and knee joints (AEDT-9810, Broadcom, 5000 CPR), at the four-bar mechanism, at the ankle joint (AMT-102, CUI, 4096 CPR), and on the treadmill (AEAT8800, Broadcom, 4096 CPR). The robot's forward motion was measured with two linear potentiometers on the four-bar slider (LX-PA-50, WayCon). The power consumption of the right and left hip and knee motors were measured individually by current sensors (ACHS-7121, Broadcom) connected to each motor's separate motor driver boards. All sensors were sampled around \SI{600}{Hz} by a single board computer (4B, Raspberry Pi Foundation). 
To avoid instabilities caused by a fixed control loop length, we adopted a dynamic sampling approach. Instead of sampling at a constant frequency, we synchronized our sampling with the completion of each control loop cycle, sending the next command only after the previous cycle had finished executing. \\
\subsection{Data analysis}
The collected data was trimmed to \SI{120}{\second} (120 gait cycles) during which the robot was steady-state walking. Since sampling intervals were not constant, sensor data were interpolated to \SI{1000}{Hz} sampling rate before analysis. Touch-down, the start of a gait cycle, was defined as the moment where the ankle joint angle starts to change after its approximately constant value (\SI{\sim 22}{deg} plantarflexion) during swing. We calculated an averaged gait cycle data from each measured sensor data, starting from touch-down. Butterworth digital low-pass zero phase shift filters with second order sections output type were applied \cite{Virtanen2020}. Separate Butterworth filter orders and cut-off frequencies were set depending on the data/sensor type, which are given in the Supplementary Table S1. In all result plots, we are presenting the averaged left leg data starting with the left foot touch-down. Equivalent plots for the averaged right leg data in the gait cycles starting with the right foot touch-down is available in the Supplementary Information S2. \\
The ankle joint power ($\mathcal{P}_\mathrm{A}$), and the GAS spring-tendon's contribution to the knee joint power ($\mathcal{P}_{K, GAS}$) was calculated as:
\begin{equation}
\mathcal{P}_\mathrm{A} = 
\begin{cases}
\begin{array}{l}
\omega_\mathrm{A} \cdot (\mathrm{k}_\mathrm{SOL} \cdot \mathrm{r}_\mathrm{SOL} \cdot (\alpha_\mathrm{A} \cdot \mathrm{r}_\mathrm{SOL} - \mathrm{l}_\mathrm{SOL, slack}) +... \\  \mathrm{k}_\mathrm{GAS} \cdot \mathrm{r}_\mathrm{GAS} \cdot ((\alpha_\mathrm{A} - \alpha_\mathrm{K}) \cdot \mathrm{r}_\mathrm{GAS} - \mathrm{l}_\mathrm{GAS, slack})),
\vspace{0.5cm}
\end{array} & \text{if} \quad \alpha_A \geq \alpha_K\\
\begin{array}{l}
\omega_\mathrm{A} \cdot \mathrm{k}_\mathrm{SOL} \cdot \mathrm{r}_\mathrm{SOL} \cdot (\alpha_\mathrm{A} \cdot \mathrm{r}_\mathrm{SOL} - \mathrm{l}_\mathrm{SOL, slack}),
\end{array} & \text{else}
\end{cases}
\end{equation}
\begin{equation}
    \mathcal{P}_\mathrm{K, GAS} = \omega_\mathrm{K} \cdot \mathrm{k}_\mathrm{GAS} \cdot \mathrm{r}_\mathrm{GAS} \cdot ((\alpha_\mathrm{A} - \alpha_\mathrm{K}) \cdot \mathrm{r}_\mathrm{GAS} - \mathrm{l}_\mathrm{GAS, slack}),
\end{equation}
where $\omega_\mathrm{A}$ is the angular velocity of the ankle in \SI{}{rad/s}, $\omega_K$ is the angular velocity of the knee in \SI{}{rad/s}, $\alpha_\mathrm{A}$ and $\alpha_\mathrm{K}$ are angle values of the ankle and knee joints in \SI{}{radian}, $\mathrm{k}_\mathrm{SOL}$ and $\mathrm{k}_\mathrm{GAS}$ are spring stiffness values of SOL and GAS respectively in \SI{}{N/m}, $\mathrm{r}_\mathrm{SOL}$ and $\mathrm{r}_\mathrm{GAS}$ are pulley radii of SOL and GAS respectively in \SI{}{meter}, $\mathrm{l}_\mathrm{SOL, slack}$ and $\mathrm{l}_\mathrm{GAS, slack}$ are the slack lengths of SOL and GAS spring-tendons respectively in \SI{}{meter}. \\
We analyzed the step-to-step transition to see the effects of both the push-off of the trailing leg and the touch-down of the leading leg. Step-to-step transition was defined from minimum to maximum vertical center of mass (CoM) velocity \cite{Adamczyk2009}.
In the velocity plots, known as 'CoM hodographs' \cite{Adamczyk2009}, from our experiments (see Supplementary Figure S1), we consistently observe two similar height vertical maximum peaks following the minimum CoM vertical velocity. We selected the second CoM vertical velocity peak for analysis, as it marks the point where the velocity vector is redirected not only vertically but also horizontally (see Supplementary Figure S1), indicating the completion of the transition \cite{Adamczyk2009}. \\
The following gait events were identified during post-processing: Start of Ankle Plantar Flexion (SAPF), Toe-Off (TO), Leading Leg Touch-Down (LLTD), Start of Hip Flexion (SHF), and Start of Knee Flexion (SKF) (\Cref{fig:joint_angles}). \\
Start of Ankle Plantar Flexion (SAPF) occurs when the trailing leg's ankle starts to plantarflex after steadily dorsiflexing during stance. We defined the moment of SAPF when the trailing leg's ankle plantar flexion angle reached its maximum. \\
Toe-Off (TO) of the trailing leg occurs when the trailing leg's foot leaves the ground. After TO, the ankle angle is approximately constant in swing (\Cref{fig:joint_angles}); at the time of TO, the ankle dorsiflexion angle reaches its minimum and then stops to change. We calculated the gradient \cite{NumPy} of the ankle angle, and defined the moment of TO when the smallest ankle dorsiflexion angle occurred within a \SI{40}{ms} time window after the largest negative peak in the ankle angle gradient. \\
Leading Leg Touch-Down (LLTD) is identified when the foot of the leading leg makes contact with the ground. Since the ankle angle remains relatively constant during the swing phase, LLTD can be detected by monitoring changes in the leading leg ankle angle. We calculated the gradient of the leading leg ankle angle \cite{NumPy}, and defined LLTD as the last instance before TO when the absolute value of the instantaneous ankle angle gradient exceeded \SI{1}{rad/s}, indicating the onset of plantarflexion or dorsiflexion. \\
Start of Hip Flexion (SHF) occurs when the hip starts flexing after steadily extending during stance. We calculated the gradient \cite{NumPy} of the hip angle, and defined the moment of SHF when the value of the instantaneous hip angle gradient was more than \SI{1}{rad/s}. The time period of interest for the search started from the moment of peak hip extension angle and lasted until the end of the gait cycle. \\
Start of Knee Flexion (SKF) occurs when the knee starts to flex during stance phase. We calculated the gradient \cite{NumPy} of the knee angle, and defined the moment of SKF when the value of the instantaneous knee angle gradient was more than \SI{1}{rad/s}. The time period of interest for the search started from \SI{0.2}{s} before SHF and lasted until \SI{0.1}{s} after SHF. \\
We considered the robot as a system of segments, divided into two feet, two shanks (lower leg), two thighs (upper leg), and a trunk (\Cref{fig:robot_design}b). We calculated the velocities at each segment's center of mass using the joint position angles from the encoders at the ankle, knee, and hip joints, at the four-bar mechanism's lower rod, and the potentiometer values that measured the horizontal position of the four-bar slider connector. Using the segments' center of mass velocity vectors ($\mathbf{v}_\mathrm{i}$), and the segments' masses ($m_\mathrm{i}$), we calculated the instantaneous linear momentum vectors of each segment ($\mathbf{p}_\mathrm{i}$): \\
\begin{equation}
    \mathbf{p}_\mathrm{i} = m \cdot \mathbf{v}_\mathrm{i},
    \label{eq:p_i}
\end{equation}
where $\mathrm{i}$ indicates one of the body segments: foot, shank, or thigh of the trailing or leading leg, or the trunk. \\
From the momentum vectors of the segments, we calculated the following measures: momentum and velocity vectors, and linear kinetic energies of the trailing leg, remaining body, and the CoM (see Supplementary Information S1).
The active joint power of the knee and hip joints could be calculated separately for each joint, as each motor was connected to a separate motor driver board and current sensor:
\begin{equation}
    P_\mathrm{joint, side} = I_\mathrm{cal, joint, side} \cdot U_\mathrm{supply},
    \label{eq:power_joint}
\end{equation}
where $P_\mathrm{joint, side}$ is the active knee/hip joint power on the trailing/leading leg side of the robot, $I_\mathrm{cal, joint, side}$ is the measured net current, at the motor driver board of the knee/hip joint on the trailing/leading leg side of the robot, and $U_\mathrm{supply}$ is the supply voltage of the robot, \SI{24}{V}. For one gait cycle, the net positive energy consumption of the two knee and two hip motors was calculated as the average of the sum of the knee and hip motor powers, after subtracting the idle power consumption of the motor drivers, and removing the negative power values. The robot's net positive cost of transport ($\mathrm{COT}$) was calculated as $\mathrm{COT}=\mathrm{E}_\mathrm{en}/(mgv)$, with $\mathrm{E}_\mathrm{en}$ as the net positive energy consumption of the knee and hip motors \cite{Seok.2013,Sprowitz.2013}, $m$ as the robot's mass, $g$ as the gravitational acceleration (\SI{9.81}{m/s^2}), and $v$ as the average horizontal robot speed over \num{120} gait cycles. Comparing the EcoWalker-2 robot's net positive cost of transport ($\coten$) in a fair manner to that of other legged hoppers, bipedal robots, and quadrupedal robots of different body mass can be achieved by calculating the robot's relative cost of transport ($\cotre$) \cite{birdbot_2022}. Hence, we calculate the ratio between the robot’s net positive cost of transport and the cost of transport of a natural runner ($\cotnr$) of equal body mass $\cotre=\coten/\cotnr \cdot \SI{100}{\%}$ \cite{tucker_energetic_1970}. \\
We used Wilcoxon signed-rank test \cite{Virtanen2020} to test the differences between the \akf{} and \pkf{} experiments for significance. In each experiment, we collected data from 120 gait cycles, providing a sample size of $\mathrm{N} = 120$ for the statistical significance tests. We used the nonparametric, paired-samples Wilcoxon test because the samples did not show normal distribution, and the samples were measured on the same robotic system, so the measurements are dependent \cite{Nehmzow2006}. We tested the time of the start of knee flexion in \akf{} and \pkf{} experiments, time of the start of ankle plantar flexion in \akf{} and \pkf{} experiments, the time period between start of ankle plantar flexion and leading leg touch-down in \akf{} and \pkf{} experiments, the absolute, horizontal, and vertical momentum change (impulse) of the trailing leg during the step-to-step transition in \akf{} and \pkf{} experiments, the absolute, horizontal, and vertical momentum change (impulse) of the remaining body during the step-to-step transition in \akf{} and \pkf{} experiments, and the absolute, horizontal, and vertical momentum change (impulse) of the CoM during the step-to-step transition in \akf{} and \pkf{} experiments. \\

%% file: tables/table_CPG_pm.tex
\begin{table}[b]
\begin{center}
\begin{tabular}{llc}
Parameter 					& 		 	                        &  Value	 \\				
\hline
Frequency                   & $f$		                        &  \SI{1}{Hz}  \\
Hip duty factor 			& $F_\text{hip}$		            &  \SI{0.6}{\%GC} \\
Knee duty factor 			& $F_\text{knee}$		            &  \SI{0.6}{\%GC} \\
Knee amplitude 				& $\Theta_\text{kneeAmplitude}$		& \SI{55}{\degree} \\
Knee offset 				& $\Theta_\text{kneeOffset}$		& \SI{8}{\degree} \\
Hip amplitude 				& $\Theta_\text{hipAmplitude}$ 		& \SI{26}{\degree} \\
Hip offset 					& $\Theta_\text{hipOffset}$			& \SI{12}{\degree} \\
Hip swing steady            & $\varphi_\text{hip}$			    & \SI{0.05}{\%GC}  \\
\end{tabular}
\end{center}
\caption{CPG control parameters for all experiments. GC: gait cycle.}
\label{tab:cpg_pm}
\end{table}